\documentclass[11pt,letterpaper]{article}
\usepackage[margin=1in]{geometry}
\usepackage{times}
\usepackage{amsmath,amssymb,bm}
\newcommand{\vect}[1]{\bm{#1}}
\usepackage{graphicx}
\usepackage{placeins}
\usepackage{booktabs}
\usepackage{array}
\usepackage{capt-of}
\usepackage[authoryear,round]{natbib}
\setcitestyle{citesep={;},aysep={,},yysep={;}}
\usepackage{url}
\usepackage[hidelinks]{hyperref}
\newcolumntype{P}[1]{>{\raggedright\arraybackslash}p{#1}}

\title{From Parameters to Answers: How LLMs Retrieve and Use Their Internal Knowledge}
\newcommand{\authorblockwidth}{0.31\textwidth}
\newcommand{\authorblock}[4]{
  \begin{minipage}[t]{\authorblockwidth}\centering
    {\normalsize #1}\par\vspace{3pt}
    {\small #2\par #3\par\href{mailto:#4}{\mbox{#4}}}
  \end{minipage}}
\newcommand{\authorlayout}{\begin{minipage}{\dimexpr\textwidth-2\tabcolsep\relax}\centering
  \authorblock{Wenkang Wei}{University of Science and\\Technology of China}{Hefei, China}{yizhilouyi@mail.ustc.edu.cn}\hfill
  \authorblock{Yuan Fang}{Singapore Management\\University}{Singapore, Singapore}{yfang@smu.edu.sg}\hfill
  \authorblock{Renhe Jiang}{The University of Tokyo}{Tokyo, Japan}{jiangrh@csis.u-tokyo.ac.jp}
  \par\vspace{16pt}
  \renewcommand{\authorblockwidth}{0.43\textwidth}
  \authorblock{Hong Cheng}{The Chinese University of Hong Kong}{Hong Kong, Hong Kong SAR, China}{hcheng@se.cuhk.edu.hk}\hspace{0.07\textwidth}
  \authorblock{Xingtong Yu}{The Chinese University of Hong Kong}{Hong Kong, Hong Kong SAR, China}{xtyu@se.cuhk.edu.hk}
\end{minipage}}
\author{\authorlayout}
\date{}
\hypersetup{pdftitle={From Parameters to Answers: How LLMs Retrieve and Use Their Internal Knowledge},pdfauthor={Wenkang Wei, Yuan Fang, Renhe Jiang, Hong Cheng, Xingtong Yu}}
\begin{document}
\maketitle
\begin{abstract}
How does a language model's dependence on query-routing information and target knowledge change as it answers a question? We study this question through layerwise interventions on the hidden state at the end of the question. Across Qwen, Llama, and Gemma, we compare country--continent questions with noun, adjective, and code answers while keeping several fitted measurements distinct. A pair-conditioned request direction describes which country is queried in natural single-country questions; a global request direction describes first- versus second-country requests in paired questions; separate selection candidates test control among contents already available in the hidden state. A diagnostic reanalysis of frozen Qwen natural-question states shows that the pair-conditioned direction grows stronger before interventions on it begin to alter later fitted knowledge, with this causal window opening while answer-supporting content is still forming. The paired three-model trajectories are not uniform: Gemma shows a partially overlapping mid-layer routing--content profile, whereas Llama has no sustained routing-effect window under the same gates. In the paired protocol, dependence on the global request direction decreases from fixed earlier to later layer sets while dependence on fitted content persists. A matched Qwen comparison shows that the pair-conditioned direction retains a late effect, so this operational handoff concerns the global fitted direction rather than all request information. These results separate early readability, natural strength, causal steering, and later content dependence.
\end{abstract}

\section{Introduction}
Large language models (LLMs) \citep{bai2023qwen,brown2020gpt3} acquire extensive knowledge during pretraining and draw on it to answer questions. Yet how a query progressively accesses this internal knowledge and turns it into an answer across layers remains unclear. Prior work has investigated where knowledge is stored in model parameters \citep{geva2021ffn,meng2022rome,meng2022memit} and identified layers or internal representations involved in its recall \citep{geva2023dissecting,yu2023mechanisms,feng2024binding}. These studies provide important but largely separate views of knowledge storage and recall. Although Geva et al. \citep{geva2023dissecting} reveal the continuous computation that connects a query to the knowledge ultimately used for an answer, none of these studies separates the information specifying which knowledge to form from the knowledge that is ultimately formed, or measures how the answer's dependence shifts between the two across layers. We therefore ask: \emph{how do LLMs progressively retrieve and use their internal knowledge across layers?}

Our analysis first reveals \textit{a functional separation between content and routing in the hidden state}. A hidden state can contribute to the answer by carrying the knowledge used to produce it or by controlling where subsequent layers access knowledge. Through controlled interventions, we causally distinguish these two roles, which we call \textbf{content} and \textbf{routing}, respectively. Routing takes two functionally distinct forms, both represented in the hidden state but directed toward different knowledge substrates: \textbf{parameter routing} points to relevant knowledge in the model's parameters, whereas \textbf{hidden-state routing} points to retrieved content already represented in the hidden state.

Tracing these components across layers further reveals \textit{how answer-supporting content is progressively formed}. In the early layers, hidden states progressively integrate query-relevant knowledge into answer-supporting content. At this stage, neither parameter routing nor hidden-state routing yet has a detectable effect on the answer. Parameter routing first grows stronger and, while answer-supporting content is still forming, begins to direct subsequent computation toward relevant knowledge in the model's parameters. As this knowledge is retrieved, it is progressively consolidated into answer-supporting content within the hidden state. Hidden-state routing becomes effective later, directing subsequent computation toward this newly formed content so that it can be further processed toward the answer.

Our analysis further reveals \textit{a routing--content handoff after answer-supporting content has formed}. We redirect parameter routing toward content that supports a specific incorrect answer and compare this intervention with random changes. This intervention generally changes the output when applied at earlier layers, but becomes less effective at later layers, where the model increasingly retains the correct answer. In contrast, intervening on the answer-supporting content remains consequential, showing that the reduced dependence on routing does not reflect the loss of this content. Instead, the remaining computation shifts its causal dependence from the routing used to locate the content toward the content itself.

In summary, our contributions are threefold. (1) We causally separate content from routing and identify parameter and hidden-state routing. (2) We reveal their ordered emergence as answer-supporting content forms across layers. (3) We identify a routing--content handoff in which causal control shifts from routing to content.

\section{Related Work}

\paragraph{Parametric knowledge representation.}
Prior work studies how language models encode knowledge acquired during pretraining. Researchers characterize feed-forward layers as associative memories and develop editing methods that locate and modify specific knowledge associations without retraining the entire model \citep{geva2021ffn,meng2022rome,meng2022memit}. These studies connect stored knowledge to model parameters, but they do not explain how a query accesses that knowledge during an ordinary forward pass. Understanding where knowledge resides therefore provides only the starting point for explaining how stored knowledge becomes an answer.

\paragraph{Internal computation of knowledge retrieval.}
Another line of work examines how language models retrieve and use pretrained knowledge during inference. These studies characterize how model components transform query information into knowledge-based predictions and decompose retrieval into functional elements such as an input, an operation, and a returned value \citep{geva2023dissecting,hernandez2024linearity,wang2025functional}. They reveal important components and stages of knowledge retrieval, but do not provide a unified account of how the model progressively transforms stored knowledge into answer-supporting content across layers.

\paragraph{Causal analysis of internal knowledge use.}
Mechanistic studies intervene on hidden states to identify internal information that influences knowledge-based predictions and trace its propagation through the model \citep{wang2024relational,hochman2026factual}. These interventions reveal where the computation causally depends on a hidden state, but an effect on the answer does not by itself identify the role that the affected information plays. The state may carry answer-supporting content or guide subsequent layers toward that content. Our work distinguishes these roles and traces how their causal contributions change across layers.

\section{Parameter-Retrieval Routing in Natural Questions}
\label{sec:parameter-routing}

We ask whether information specifying a requested fact helps the model form that fact in later computation. To test this, we ask country--continent questions and compare the model's internal states when the requested country changes. We separate a request-related direction from a fitted measure of continent content, change that direction, and measure both the later content and the answer. The results distinguish a request that is readable early from a request component that becomes stronger and later affects computation: reversal changes both the answer and later fitted knowledge while the content is still developing.

\subsection{Changing the Country Changes the Requested Stored Fact}

Our first task is to distinguish the information specifying a requested country from the continent knowledge needed to answer. We use questions that name a country and ask for its continent. For example, the following question requires Africa:
\begin{center}
\small
\fbox{\begin{minipage}{0.83\linewidth}
\ttfamily
Which continent is Kenya located in?\\
Answer with only the continent name.
\end{minipage}}
\end{center}
Changing Kenya to China asks for Asia instead. Appendix~\ref{app:natural-inputs} lists every fitting country, screening pair, validation pair, and question wording.
Because the answer must come from the model's stored association, each prompt gives the country and the relation to be answered, but not the continent. The model processes each question separately. We then compare the recorded states of these two computations; the two countries form an analysis pair, not a joint model input. Africa and Asia label the requested facts in our measurements and candidate-answer comparisons, rather than supplying those facts to the question.

We test whether this country distinction survives a change in wording. Let $s$ denote a country and $\tau\in\{A,B,C\}$ a question wording. The text $p_{\tau,s}$ is the question and answer instruction obtained by inserting $s$ into that wording. This question template is distinct from the checkpoint's \emph{chat template}, which wraps the text in user/assistant role markers. We denote the resulting tokenized model input by $x_{\tau,s}$; its construction is defined below. Template A is the wording above. Template B asks \emph{On which continent is Kenya located?}, and template C asks \emph{What continent is Kenya in?}, with the country substituted as appropriate. All three ask for only the continent name. We use template A to construct the state comparison and B and C to evaluate whether that comparison transfers to other wordings.

Separate country sets keep the construction and evaluation distinct. The fitting set is
$\mathcal{C}_{\mathrm{fit}}=\{\mathrm{China},\mathrm{France},\mathrm{Kenya},\mathrm{Nigeria},\mathrm{Spain},\mathrm{Thailand}\}$,
with two countries from each of Africa, Asia, and Europe. These six countries provide the reference states used to measure continent content. Three further, disjoint pairs (six countries and 12 questions under B and C) screen the intervention layers, and 24 additional pairs validate the frozen choice. Each pair contains countries from different continents, so changing the country changes the requested fact. A validation pair supplies four questions---its two countries under B and C---but remains one independent statistical unit.

Changing a country also changes its name and tokenization. A difference between the two recorded states therefore does not by itself identify retrieval control. The next subsection defines which part of this difference is measured; the subsequent interventions test whether that part affects later knowledge and the answer.

\subsection{Separating the Request Difference from Continent Content}

To test a request-related state change, we first specify where the state is taken and how we distinguish the requested fact from the request. We use Qwen-2.5-3B-Instruct \citep{qwen2024qwen25}, a decoder-only language model with $L=36$ decoder blocks and no separate encoder stack. A decoder block is one successive processing layer: causal self-attention combines information from the current and preceding input positions, and a feed-forward network transforms the resulting representations \citep{vaswani2017attention}. The blocks update a numerical vector at every token position before the model predicts the next token.

To locate the vector consistently, we follow how the question becomes a model input. A tokenizer splits text into the units processed by the model; subword tokenization allows a word to occupy more than one unit \citep{sennrich2016subword}. The fixed checkpoint first applies its chat formatting and then tokenization:
\[
x_{\tau,s}=\mathtt{Tokenize}\!\left(\mathtt{Chat}(p_{\tau,s})\right).
\]
Here $\mathtt{Chat}$ inserts the checkpoint's role and generation markers, and $\mathtt{Tokenize}$ returns their token sequence together with the question tokens. Both operations are fixed by the released checkpoint. Write an input as $x=(x_1,\ldots,x_{n_x})$, where $n_x$ is its token count after chat formatting. The \emph{final input token} is position $n_x$, immediately before the answer continuation; it need not be the last ordinary word of the question. We call this the \emph{question-end position}. Bold symbols denote vectors or matrices, and layer indices appear in superscripts. Let $\vect{H}_{x}^{\ell}\in\mathbb{R}^{n_x\times H}$ collect all position vectors after decoder block $\ell$, where $H$ is the vector dimension. The block and the vector we record can be written as
\[
\begin{aligned}
\vect{H}_x^{0}&=\mathtt{Embed}(x;\theta_{\mathrm{emb}}),\\
\vect{H}_{x}^{\ell}&=\mathtt{Block}^{\ell}(\vect{H}_{x}^{\ell-1};\theta^{\ell}),
\qquad \ell\in\{1,\ldots,L\},\\
\vect{h}_{x}^{\ell}&=\vect{H}_{x}^{\ell}[n_x,:]^\top.
\end{aligned}
\]
Here, $\mathtt{Embed}$ looks up each input token's embedding using the fixed embedding parameters $\theta_{\mathrm{emb}}$, so $\vect{H}_x^{0}$ contains one numerical vector per input position; $\theta^{\ell}$ denotes the fixed parameters of block $\ell$, and $\mathtt{Block}^{\ell}$ denotes the model's actual layer computation with the positions and causal mask fixed by $x$. Thus, $\vect{h}_{x}^{\ell}\in\mathbb{R}^{H}$ is one recorded question-end vector, not the whole sequence or the model's parameters.

We use this position because its later computation leads to the next-token prediction. \emph{Recording} copies its values for analysis without changing the model. \emph{Intervening} instead changes that vector at a chosen block and lets the remaining blocks compute from it; the next subsection specifies the changes. Both operations refer to decoder-block outputs before the model's final normalization, the rescaling applied after the last block and before output scoring. Qwen uses root mean square layer normalization for this operation \citep{qwen2024qwen25,zhang2019rmsnorm}. We use an input index $x$ throughout: the A in $x_{A,s}$ specifies the input wording, not an additional intervention condition.

We next build a reference for the continent information in these vectors. Let $\mathcal{K}=\{\mathrm{Africa},\mathrm{Asia},\mathrm{Europe}\}$ be the set containing the three continent labels. For $k\in\mathcal{K}$, $\mathcal{C}_{\mathrm{fit},k}$ contains the two fitting countries with label $k$. Throughout this construction the wording is fixed to A; write $x_s=x_{A,s}$ and record one question-end vector for each fitting country. The overall and continent-specific means are
\[
\vect{\mu}^{\ell}=\frac{1}{6}\sum_{s\in\mathcal{C}_{\mathrm{fit}}}\vect{h}_{x_s}^{\ell},
\qquad
\vect{\mu}_{k}^{\ell}=\frac{1}{2}\sum_{s\in\mathcal{C}_{\mathrm{fit},k}}\vect{h}_{x_s}^{\ell}.
\]
These are elementwise averages of hidden-state vectors at the same layer and position, not averages of country-name embeddings, all input tokens, or all three wordings. They are fitted once per layer and shared across the evaluated questions, which is why they carry no evaluation-input index.

The three centered continent means define the directions needed to measure differences among the requested facts. Let $k_1,k_2,k_3$ be the labels in the displayed order of $\mathcal{K}$. Concatenating their mean-difference columns and transposing gives
\begin{equation}
\vect{V}^{\ell}
=\bigl[\vect{\mu}_{k_1}^{\ell}-\vect{\mu}^{\ell},\
       \vect{\mu}_{k_2}^{\ell}-\vect{\mu}^{\ell},\
       \vect{\mu}_{k_3}^{\ell}-\vect{\mu}^{\ell}\bigr]^\top
\in\mathbb{R}^{3\times H}.
\label{eq:single-query-knowledge-fit}
\end{equation}
We obtain an orthonormal basis for these differences using the singular value decomposition \citep[Section~3.2.2]{halko2011structure}:
\[
\vect{U}^{\ell}=\mathtt{RightSV}(\vect{V}^{\ell};2),
\qquad
\vect{P}^{\ell}=\vect{U}^{\ell}(\vect{U}^{\ell})^{\top}.
\]
The operation $\mathtt{RightSV}(\vect{V};2)$ decomposes its input matrix and returns the two right singular vectors associated with its largest singular values as columns. Thus $\vect{U}^{\ell}\in\mathbb{R}^{H\times2}$ contains two perpendicular unit axes, and $\vect{P}^{\ell}\in\mathbb{R}^{H\times H}$ is their orthogonal projection matrix \citep[Section~3.1]{halko2011structure}. The matrix is not itself a recorded fact or a named model operation: multiplication by it projects a vector onto the fitted axes. This supplies a two-dimensional content measure, rather than an exhaustive representation of the model's continent knowledge.

Before testing the remaining request difference, we also distinguish it from choosing between two supplied records. A separate calibration task supplies records such as \emph{Kenya has marker dax} and \emph{China has marker fep}, then requests the first or second marker. Its answers are present in the input, so it calibrates record selection without requiring a country--continent association. At each layer, we subtract the mean recorded vector for second-record requests from the mean for first-record requests in the calibration fit set, remove its projection through $\vect{P}^{\ell}$, and normalize the remainder. Denote the resulting unit vector by $\vect{\eta}^{\ell}$. Section~\ref{sec:object-selection} describes the calibration test. Excluding this specific contrast prevents the candidate below from simply reusing that fitted selection direction.

For a country pair $(s,s')$, we now remove the fitted content projection from the difference between its two template-A states:
\begin{equation}
\widetilde{\vect{d}}^{\ell}(s,s')
=
(\vect{I}_{H}-\vect{P}^{\ell})
\bigl(\vect{h}_{x_s}^{\ell}-\vect{h}_{x_{s'}}^{\ell}\bigr),
\label{eq:single-query-knowledge-exclusion}
\end{equation}
where $\vect{I}_{H}$ is the $H$-dimensional identity matrix and the tilde marks the content-excluded difference. We then remove its component along the calibrated selection direction and normalize:
\begin{equation}
\vect{r}^{\ell}(s,s')
=
\mathtt{Unit}\!\left[
\widetilde{\vect{d}}^{\ell}(s,s')
-\bigl((\vect{\eta}^{\ell})^{\top}\widetilde{\vect{d}}^{\ell}(s,s')\bigr)\vect{\eta}^{\ell}
\right].
\label{eq:single-query-route}
\end{equation}
For nonzero $\vect{v}$, $\mathtt{Unit}(\vect{v})=\vect{v}/\|\vect{v}\|_2$. The direction $\vect{r}^{\ell}(s,s')$ retains the part of the country-request difference outside these two fitted components. We call it a \emph{parameter-retrieval-routing candidate}. It points from the second member toward the first in this fitted comparison; whether it guides knowledge formation is the question tested next.

\subsection{Deleting and Reversing the Request Component}

We test the candidate by changing only the recorded state's coordinate along it. Deletion brings that coordinate to the midpoint between the two requests; reversal moves it to the opposite request's side. Comparing their downstream effects with equally large random changes tests whether the candidate matters beyond generic disruption of a state.

For a held-out template $\tau\in\mathcal{T}_{\mathrm{test}}=\{B,C\}$ and pair $(s,s')$, write $x=x_{\tau,s}$ and $\bar{x}=x_{\tau,s'}$. These are two inputs with the same wording and different requested countries. Their midpoint is
\begin{equation}
\vect{b}_{x}^{\ell,\mathrm{pair}}
=\tfrac12\bigl(\vect{h}_{x}^{\ell}+\vect{h}_{\bar{x}}^{\ell}\bigr).
\end{equation}
The superscript $\mathrm{pair}$ identifies a center computed from this pair, not from the fitting set. The coordinate of either input along the fixed template-A direction is
\begin{equation}
\alpha_{x}^{\ell}
=
\vect{r}^{\ell}(s,s')^{\top}
\bigl(\vect{h}_{x}^{\ell}-\vect{b}_{x}^{\ell,\mathrm{pair}}\bigr).
\label{eq:single-query-coefficient}
\end{equation}
This scalar is a signed projection: the paired input has the opposite coefficient because both states use the same midpoint. Deletion and reversal produce the following edited vectors:
\begin{equation}
\vect{h}_{x}^{\ell,\mathrm{delete}}
=
\vect{h}_{x}^{\ell}-\alpha_{x}^{\ell}\vect{r}^{\ell}(s,s'),
\label{eq:single-query-delete}
\end{equation}
\begin{equation}
\vect{h}_{x}^{\ell,\mathrm{flip}}
=
\vect{h}_{x}^{\ell}-2\alpha_{x}^{\ell}\vect{r}^{\ell}(s,s').
\label{eq:single-query-flip}
\end{equation}
The superscript $\ell$ identifies the source layer; the labels $\mathrm{delete}$ and $\mathrm{flip}$ identify the two intervention operations. We write the edited vector only at the source block's question-end position. The input, model weights, and all other positions at that block stay unchanged. The remaining blocks then compute new states and answer probabilities; these downstream changes, rather than the written vector itself, are our outcomes.

The comparison requires the intended edit to survive numerical rounding. Some pair-conditioned changes round to zero in the 16-bit floating-point format bfloat16. We therefore evaluate the frozen weights using the 32-bit floating-point format float32 and verify the vector actually written. All conditions in this section use float32. The later paired-country protocol uses bfloat16; comparisons across these protocols concern qualitative patterns and within-protocol contrasts, not raw effect magnitudes.

For each true change $\Delta\vect{h}_{x}^{\ell,c}=\vect{h}_{x}^{\ell,c}-\vect{h}_{x}^{\ell}$, with $c\in\{\mathrm{delete},\mathrm{flip}\}$, we generate eight independent Gaussian vectors $\vect{\epsilon}_{x}^{\ell,c,j}\sim\mathcal{N}(\vect{0},\vect{I}_{H})$. The index $j\in\{1,\ldots,8\}$ identifies a random control. We match each random change to the true intervention's length:
\begin{equation}
\Delta\vect{h}_{x}^{\ell,c,j}
=
\left\|\Delta\vect{h}_{x}^{\ell,c}\right\|_2
\frac{\vect{\epsilon}_{x}^{\ell,c,j}}
{\left\|\vect{\epsilon}_{x}^{\ell,c,j}\right\|_2},
\qquad j\in\{1,\ldots,8\}.
\label{eq:single-query-random}
\end{equation}
Length matching uses the actual float32 write, and deletion and reversal have separate controls. We also check the ratio
$\|\vect{P}^{\ell}\Delta\vect{h}_{x}^{\ell,c}\|_2/\|\Delta\vect{h}_{x}^{\ell,c}\|_2$:
the size of any change directly reintroduced into the fitted continent space by rounding, divided by the full change length. These checks verify that the tested numerical operation preserves its stated size and fitted-content exclusion.

\subsection{Measuring Answer Changes and Later Fact Changes}

The intervention can change the answer without changing the particular fact representation we measure. We therefore use two outcomes: the model's preference between the requested and paired answers, and the later state's proximity to the two fitted continent references. A separate, non-interventional measure asks whether the unmodified request distinction is readable. Keeping these measures separate distinguishes information present from information with a downstream effect.

For question $x$, let $a_x^+$ be the requested country's continent name and $a_x^-$ the paired country's continent name. A Kenya--China pair therefore compares Africa with Asia. These are evaluator-provided answer continuations, not facts included in the question. For candidate $a=(a_1,\ldots,a_{T_a})$, where $T_a$ is its token count, define
\begin{equation}
\lambda_{x,a}^{e}
=
\frac{1}{T_a}\sum_{t=1}^{T_a}
\log\mathtt{Prob}(a_t\mid x,a_{<t},e;\theta),
\qquad
m_x^e=\lambda_{x,a_x^+}^{e}-\lambda_{x,a_x^-}^{e}.
\label{eq:single-query-answer-margin}
\end{equation}
Here, $\mathtt{Prob}$ is the model's next-token probability, $\theta$ its fixed parameters, and $a_{<t}$ the preceding tokens of the candidate being scored. The condition $e$ specifies no intervention ($e=0$), a true edit, or a random control; $x$ remains the input in every condition. Thus $\lambda_{x,a}^{e}$ is a length-averaged log probability and $m_x^e$ the correct-minus-paired answer margin. We score the two fixed continuations rather than unrestricted generation. The loss $\Delta m_x^e=m_x^0-m_x^e$ is positive when the correct answer loses some of its advantage.

To determine whether the targeted edit causes more loss than random disruption, let $\mathcal{X}$ be the evaluated questions and $N=|\mathcal{X}|$ their number. For a source layer $\ell$, write $e=(\ell,c)$ for true intervention $c$ and $e=(\ell,c,j)$ for its random control $j$. Accordingly, $\Delta m_{x}^{\ell,c}\equiv\Delta m_x^{(\ell,c)}$ and $\Delta m_{x}^{\ell,c,j}\equiv\Delta m_x^{(\ell,c,j)}$. The corrected answer effect is
\begin{equation}
E_{c}^{\ell,\mathrm{ans}}
=
\frac1N\sum_{x\in\mathcal{X}}\Delta m_{x}^{\ell,c}
-\max_{j=1,\ldots,8}
\frac1N\sum_{x\in\mathcal{X}}\Delta m_{x}^{\ell,c,j}.
\label{eq:single-query-answer-effect}
\end{equation}
The layer index in the superscript identifies where the edit was made, and $\mathrm{ans}$ identifies the answer outcome. We first average each condition over questions, then subtract the largest random-control average. A positive effect means greater mean margin damage than every tested equal-length random control, not an answer switch on every question.

The internal outcome uses the content space fitted at a receiving layer $\ell'>\ell$. For a fixed source intervention, let $\vect{h}_{x}^{\ell',e}$ be the receiving state in condition $e$ and project it as $\vect{z}_{x}^{\ell',e}=\vect{P}^{\ell'}(\vect{h}_{x}^{\ell',e}-\vect{\mu}^{\ell'})$, and define the reference for continent $k$ as $\vect{\kappa}_{k}^{\ell'}=\vect{P}^{\ell'}(\vect{\mu}_{k}^{\ell'}-\vect{\mu}^{\ell'})$. For the requested and paired continent labels $k_x^+$ and $k_x^-$, the knowledge score is
\begin{equation}
\gamma_{x}^{\ell',e}
=
\|\vect{z}_{x}^{\ell',e}-\vect{\kappa}_{k_x^-}^{\ell'}\|_2^2
-\|\vect{z}_{x}^{\ell',e}-\vect{\kappa}_{k_x^+}^{\ell'}\|_2^2.
\label{eq:single-query-knowledge-score}
\end{equation}
A positive score means that the projected state is closer to the correct reference. All conditions use the same fixed projector, mean, and references; only the receiving state changes. Without an intervention label, $\vect{h}_{x}^{\ell'}$, $\vect{z}_{x}^{\ell'}$, and $\gamma_{x}^{\ell'}$ denote their unmodified ($e=0$) values. With source layer $\ell$ fixed, superscripts $c$ and $(c,j)$ abbreviate its true edit and random control. The loss after source intervention $c$ is $\Delta\gamma_{x}^{\ell\rightarrow\ell',c}=\gamma_{x}^{\ell',0}-\gamma_{x}^{\ell',c}$. Replacing superscript $c$ by $(c,j)$ defines the corresponding random-control loss $\Delta\gamma_{x}^{\ell\rightarrow\ell',c,j}$. Its random-corrected effect is
\begin{equation}
E_{c}^{\ell\rightarrow\ell',\mathrm{know}}
=
\frac1N\sum_{x\in\mathcal{X}}\Delta\gamma_{x}^{\ell\rightarrow\ell',c}
-\max_{j=1,\ldots,8}
\frac1N\sum_{x\in\mathcal{X}}\Delta\gamma_{x}^{\ell\rightarrow\ell',c,j}.
\label{eq:single-query-later-knowledge}
\end{equation}
The arrow names the source and receiving layers, and $\mathrm{know}$ identifies the knowledge outcome. The aggregation matches Equation~\ref{eq:single-query-answer-effect}, but the unit is a fitted squared-distance score, not answer log probability. We therefore do not compare the two effect magnitudes.

Before evaluating new countries, the 12 screening questions---three pairs, two countries per pair, and templates B and C---fix one source and one receiver. Both deletion and reversal must have positive answer effects at the source. At the receiver, the unmodified score must favor the correct continent on at least 80\% of screening questions, and both edits must have positive later-knowledge effects. The fixed pair is then evaluated on the 24 validation country pairs. We obtain 95\% percentile bootstrap intervals \citep{efron1986bootstrap} by resampling country pairs while keeping each pair's four questions together.

For comparison with these intervention effects, we measure readability in the unmodified states. Fix the order $(s,s')$ of each pair and set $\sigma_x=+1$ when $x$ requests $s$ and $-1$ when it requests $s'$. Using the held-out-input coefficient in Equation~\ref{eq:single-query-coefficient}, define
\begin{equation}
F^{\ell,\mathrm{route}}
=\frac1N\sum_{x\in\mathcal{X}}
\mathtt{Ind}(\sigma_x\alpha_{x}^{\ell}>0).
\label{eq:single-query-pointing}
\end{equation}
The indicator $\mathtt{Ind}$ returns one if its condition holds and zero otherwise. This fraction counts questions whose coefficient points toward their requested pair member. It tests whether the pair's template-A distinction transfers to other wordings, not whether a classifier trained on old countries identifies unseen countries independently.

\subsection{Early Readability Precedes Detectable Answer Dependence}

\begin{figure}[t]
\centering
\includegraphics[width=\linewidth]{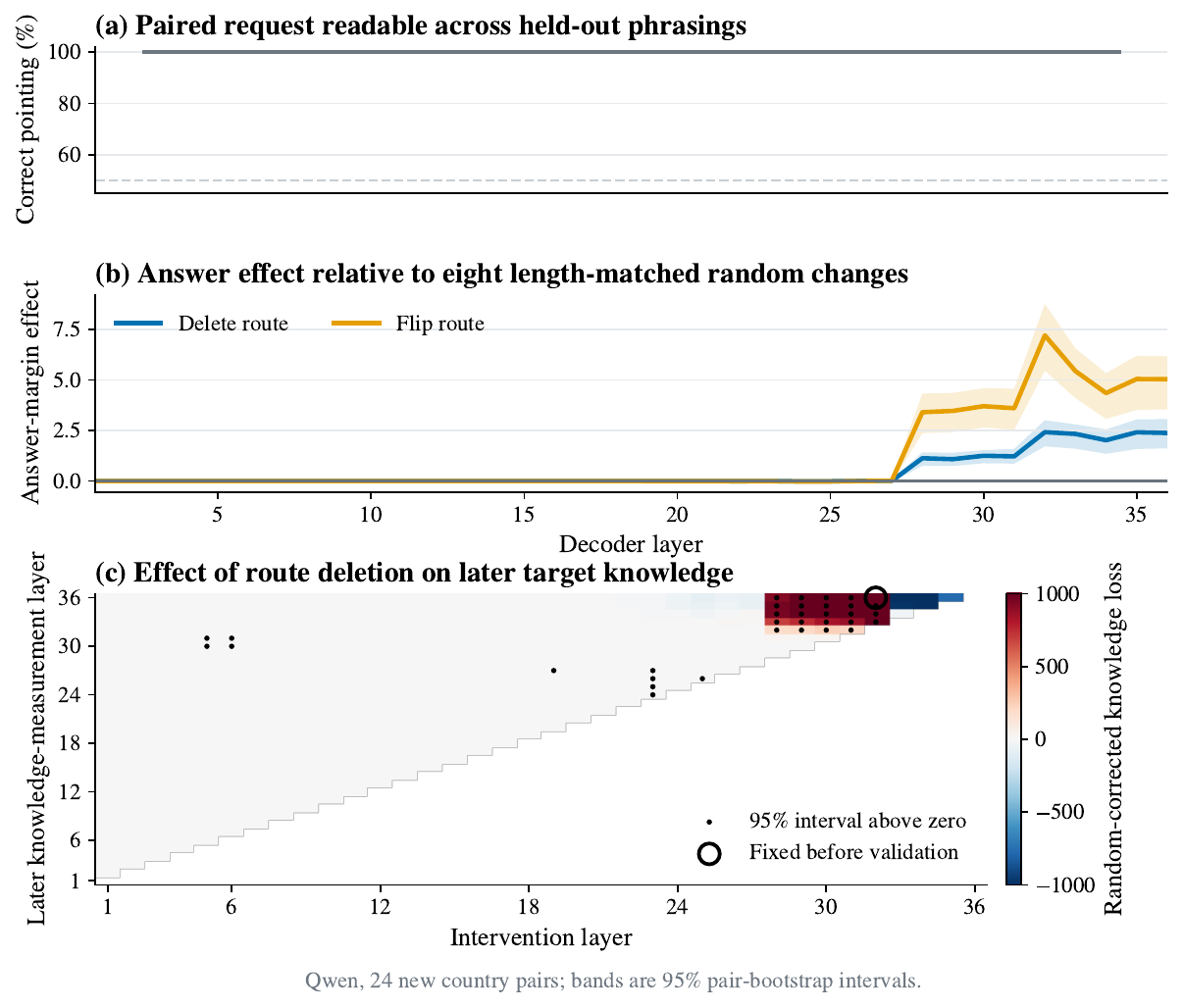}
\caption{\textbf{A readable pair-conditioned request distinction and detectable answer dependence have different depth profiles in Qwen.} Every model input is a single-country question; two countries are paired only for analysis. Template A constructs one direction per pair, and two held-out phrasings evaluate it on 24 new pairs. (a) Correct pointing is the fraction of questions for which the coefficient sign identifies the queried pair member. (b) Answer effect is the mean correct-versus-paired margin loss minus the largest mean loss among eight random changes matched to the actual intervention length. Lines show estimates; bands show 95\% country-pair bootstrap intervals. (c) Color shows the similarly corrected loss of the fitted continent score at a later layer after source-layer deletion. Black dots mark individual layer pairs whose 95\% interval lies above zero, not a simultaneous matrix test. The circle marks source layer 32 and receiver layer 36, fixed before validation. Deletion at that pair has an interval spanning zero for later knowledge, whereas reversal passes. Values use float32 and are not compared in magnitude with the bfloat16 paired-task experiments.}
\label{fig:natural-single-country-qwen}
\end{figure}

The candidate distinguishes the requested country long before its deletion or reversal measurably affects the answer. Figure~\ref{fig:natural-single-country-qwen} compares this readable distinction with the intervention effects across all 36 layers; Table~\ref{tab:natural-single-country-qwen} then tests whether the fixed source edit also changes knowledge at the fixed receiver. Qwen prefers the correct continent on all 12 screening and 96 validation questions before intervention, so these comparisons start from correct baseline answers.

The numerical and replication checks support interpreting the measured changes as the intended interventions. A fresh model reload reproduces every recorded margin, knowledge score, pointing value, and intervention-quality measurement. The failure rate is below 5\% for random-length matching, target writes, and fitted-knowledge leakage.

Figure~\ref{fig:natural-single-country-qwen} separates two questions. The top panel asks whether the pair-conditioned direction identifies which member is being queried; it succeeds on every held-out question at every layer. The middle panel asks whether removing or reversing that direction damages the answer more than matched random changes; its 95\% intervals rise above zero only at layers 28--36, with both effects peaking at layer 32. Thus, readability is present early, whereas detectable answer dependence is concentrated late.

The selection questions fix layer 32 as the intervention layer and layer 36 as the receiving layer before the 24 validation pairs are analyzed. Table~\ref{tab:natural-single-country-qwen} gives the four registered comparisons.

\begin{table}[t]
\centering
\caption{Registered Qwen comparisons at the layer pair fixed on selection questions. Each effect is the true mean loss minus the largest mean loss among eight actual-length-matched random controls. The unit is an answer-margin change in the first two rows and a fitted knowledge-score change in the last two; these units are not comparable. Intervals resample 24 country pairs.}
\label{tab:natural-single-country-qwen}
\small
\begin{tabular}{lrr}
\toprule
Intervention and measured outcome & Effect & 95\% interval\\
\midrule
Delete route; answer margin & 2.418 & [1.721, 2.976]\\
Flip route; answer margin & 7.200 & [5.422, 8.634]\\
Delete route; layer-36 knowledge & 1102.517 & [$-$62.431, 2133.854]\\
Flip route; layer-36 knowledge & 6270.843 & [3408.236, 8771.132]\\
\bottomrule
\end{tabular}
\end{table}

At the fixed layer pair, deleting the layer-32 direction reduces the answer margin beyond every tested random direction. Reversal has a larger answer effect and also reduces the layer-36 knowledge score beyond its own, longer controls. However, the deletion effect on later knowledge has an interval that crosses zero. The registered four-part criterion therefore does not pass in full: the experiment supports a reversible steering effect, but not deletion-based necessity for later knowledge formation.

A sensitivity analysis combines these controls with eight additional random controls from the common-protocol comparison in Section~\ref{sec:matched-direction-comparison}. The 96 corresponding questions have identical raw deletion effects. With all sixteen controls, the later-knowledge deletion interval still spans zero, whereas the reversal interval remains positive (Appendix~\ref{app:random-control-sensitivity}). The registered deletion boundary is therefore retained rather than replaced by a more favorable random-control sample.

These are continuous score changes, not universal answer switches. At layer 32, all 96 unmodified validation questions prefer the correct candidate. After deletion, all still do so despite their reduced margins. Reversal makes 15 of 96 prefer the paired answer. Replacing the entire hidden state with the paired question's state switches all 96, but that diagnostic transfers knowledge and other information together and is not treated as a routing-specific effect.

\subsection{Natural Route Strength Precedes Its Later-Knowledge Effect}

Readability alone leaves a timing question unresolved: does the request component strengthen before it starts to influence later knowledge, or only after the fact is already available? We compare the natural coefficient with content formation and intervention effects in the same frozen validation records. The comparison shows a sustained strength increase before the later-knowledge effect, with content still developing during that interval.

The pointing fraction in Figure~\ref{fig:natural-single-country-qwen} uses only the coefficient's sign. To measure its magnitude relative to the model state, use the question-end vectors $\vect{h}_{x}^{\ell}$ and $\vect{h}_{\bar{x}}^{\ell}$ for paired inputs $x$ and $\bar{x}$. Define the state-relative natural route strength as
\[
S_{x}^{\ell,\mathrm{route}}
=\frac{|\alpha_{x}^{\ell}|}
{\sqrt{\bigl(\|\vect{h}_{x}^{\ell}\|_2^2+\|\vect{h}_{\bar{x}}^{\ell}\|_2^2\bigr)/2}}.
\]
The numerator is the natural coefficient in Equation~\ref{eq:single-query-coefficient}; because the route direction has unit length, it is also the intended deletion length before numerical rounding. The denominator is the root-mean-square length of the two full states. Thus, $S^{\mathrm{route}}$ measures the coefficient relative to the whole-state scale; it is not a percentage of the model's information, parameters, or neurons.

We first average the two held-out phrasings and two requested countries within each of the 24 validation pairs. Layers 1--12 define an early reference separately for each pair. A descriptive onset is the first of three consecutive later layers whose pair-bootstrap 95\% interval for the increase above that reference lies above zero. This rule places the sustained state-relative increase at layer 21: the mean rises from an early reference of $0.00905$ to $0.01412$ at layer 21, $0.04963$ at layer 27, and $0.10128$ at layer 28. Pair alignment does not show a new onset, consistent with a direction that is already readable early and whose coefficient later becomes larger.

The fitted fact is still changing over this interval. The fraction of questions whose unmodified state favors the correct continent is $69.8\%$ at layer 21, $75.0\%$ at layer 24, $82.3\%$ at layer 26, and $96.9\%$ at layer 28; it reaches $100\%$ only at layer 33. Removing fitted content begins to damage the answer beyond matched random changes at layer 27. One layer later, deleting the natural route coefficient yields a positive random-corrected effect on the fixed layer-36 fact score at source layers 28--31, while reversal does so at layers 28--32. These are pointwise diagnostic intervals from an existing-data reanalysis, not simultaneous confidence bands or a prospectively registered universal onset.

\begin{figure}[t]
\centering
\includegraphics[width=\linewidth]{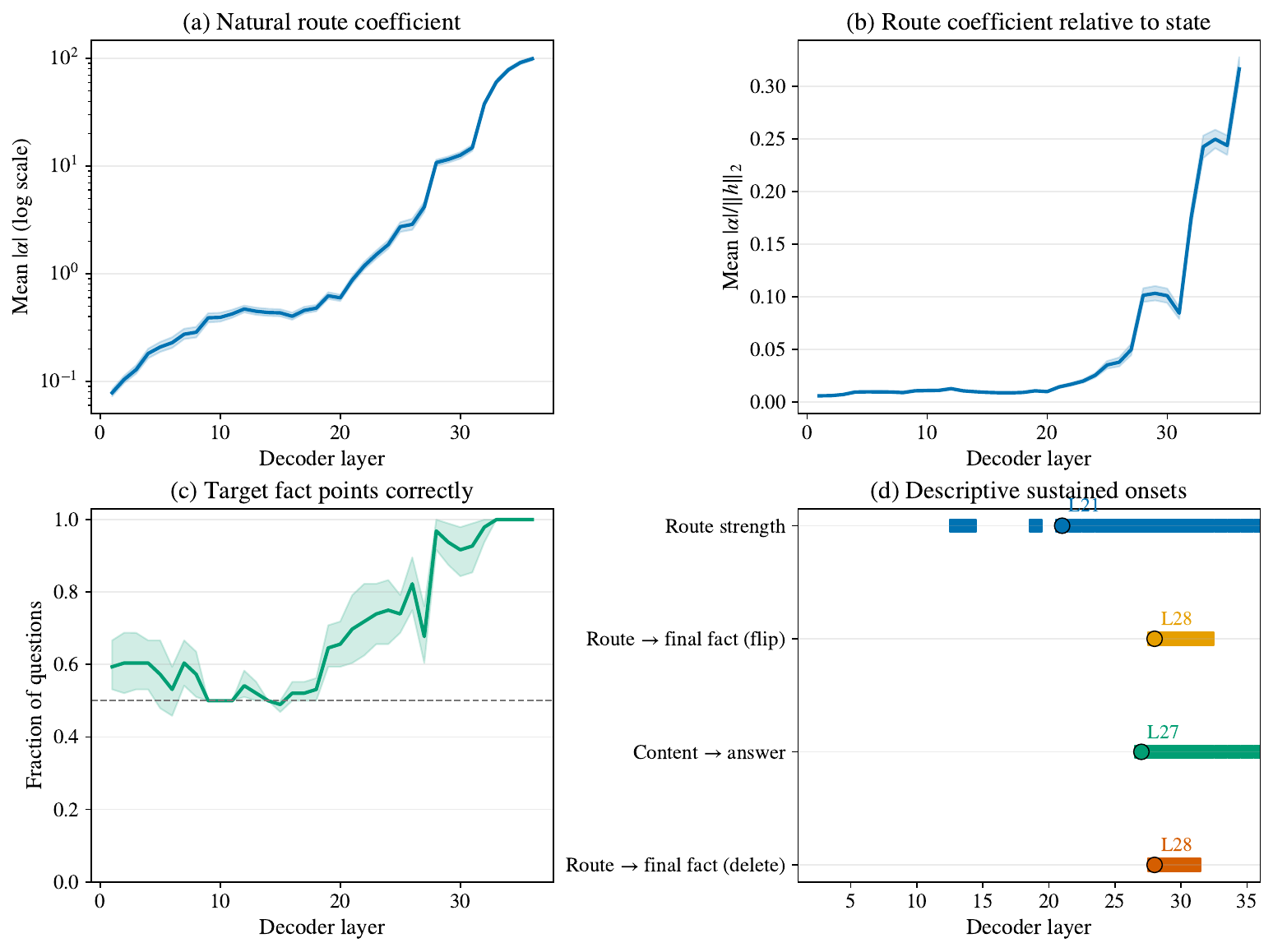}
\caption{\textbf{In Qwen natural questions, route strengthening precedes a later causal window while fitted content is still forming.} (a) The absolute natural route coefficient is shown on a logarithmic scale. (b) Dividing that coefficient by the root-mean-square full-state length removes a simple whole-state scale explanation. (c) The fraction of questions whose fitted fact score favors the correct continent rises gradually rather than appearing only after the route is complete. (d) Squares mark layers whose pointwise country-pair bootstrap interval passes the stated comparison; circles mark the first run of three consecutive passing layers. Route strength first has a three-layer run at layer 21, fitted-content deletion affects answers from layer 27, and route deletion or reversal affects the fixed layer-36 fact from layer 28. The route-to-fact windows end at layers 31 and 32 rather than continuing to the model output. This is a diagnostic reanalysis of the 24 frozen validation pairs.}
\label{fig:natural-route-strength-timing}
\end{figure}

This sequence supports an overlapping mechanism rather than a discrete pipeline. The pair-conditioned route grows stronger before its intervention has a stable effect on the final fitted fact, but fitted fact information is already developing and becomes answer-relevant during the same period. Appendix~\ref{app:route-strength-timing} asks whether the paired-country records for Qwen, Llama, and Gemma show the same schedule; they do not provide a direct cross-model replication.

\subsection{Reversal Steers Later Knowledge without Fixed-Pair Deletion Necessity}

The combined results distinguish representing a request from using its measured component to guide a fact. After the two fitted exclusions, the candidate preserves the country distinction across wordings from the first layer. Its natural coefficient strengthens later, while deletion and reversal affect answers detectably only in a late range. At the fixed source layer 32, reversal also changes the fitted knowledge score at receiving layer 36. Thus an imposed movement along the request candidate can steer later computation toward the paired country's knowledge.

Deletion provides the narrower result: it changes the answer margin but does not give a confirmed layer-36 knowledge loss. The experiment therefore establishes directional steering without establishing that the naturally occurring coefficient is necessary for that later fitted fact.

We retain the term \emph{parameter-retrieval-routing candidate} for this measured direction rather than identifying it with a unique physical address or all retrieval information. Sections~\ref{sec:joint-process} and \ref{sec:matched-direction-comparison} test whether the later dependence profile changes when the request direction is fitted globally instead of separately for each pair.

\section{Object Selection within Hidden States}
\label{sec:object-selection}

When a question names two countries but requests only one answer, does the model form just the requested fact, or keep both facts usable and select between them? For example, a question listing Kenya and China and asking for the first country's continent could be answered by retrieving only Africa. We reserve \emph{object selection} for the different function of choosing which of two internally available candidate contents controls the answer.

We test this function first when both answers are supplied, then transfer the fitted selection contrast to questions whose facts must come from the model. A direct test goes further: after changing the selected object, it asks whether two continuations from the same modified state can still produce the two candidate answers. The transfer test gives a format-specific result, while the strict two-continuation test succeeds in a separate Qwen base-model capital task and supplies the reference for the instruction-model comparisons below.

\subsection{Calibrating Selection when Both Answers Are Supplied}

To measure selection separately from recalling an unstated fact, we begin with a task that supplies both possible answers. Changing the requested record then changes only which supplied marker should be returned. The calibration prompt is:
\begin{quote}
\small
First record: Kenya has marker dax.\\
Second record: China has marker fep.\\
Which marker belongs to the requested \textbf{first} record?\\
Answer with only the marker.
\end{quote}
Changing \emph{first} to \emph{second} changes the requested marker from dax to fep, without requiring either marker to be recalled from the weights. We record the question-end vector $\vect{h}_{x}^{\ell}$ at the position defined in Section~\ref{sec:parameter-routing}. The calibration fitting set contains three record pairs under both requests, or six questions. At each layer, we average the three first-record vectors and subtract the average of the three second-record vectors. The resulting raw contrast $\vect{o}^{\ell,\mathrm{raw}}\in\mathbb{R}^{H}$ describes a choice between supplied records; the next test asks whether it also contributes when the factual answers are not supplied.

We next ask whether part of this contrast matters when the prompt supplies the objects but not their factual answers:
\begin{quote}
\small
First country: Kenya\\
Second country: China\\
Which continent is the \textbf{first} country located in?\\
Answer with only the continent name.
\end{quote}
The prompt names Kenya and China and specifies which position to answer, but it never states Africa or Asia. The model must therefore use stored country--continent knowledge. We repeat the task with continent names, adjectives such as African and Asian, and arbitrary output codes whose mappings are defined in the prompt (for example, Africa = \emph{dax}). These formats alter how the same association must be expressed.

This transfer comparison uses the text-processing decoders of Qwen-2.5-3B-Instruct \citep{qwen2024qwen25}, Llama-3.2-3B-Instruct \citep{dubey2024llama3}, and Gemma-3-4B-Instruct \citep{gemmateam2025gemma3}, with 36, 28, and 34 decoder blocks, respectively, in bfloat16. The inputs contain text only. The state dimension $H$ is specific to each model.

The paired-country task has its own fitted means and continent projector $\vect{P}^{\ell}$, constructed by the centered-continent-mean procedure in Equation~\ref{eq:single-query-knowledge-fit}, but from its own fitting questions and separately for each answer format. These quantities are fixed within a model--format comparison, not shared with the natural single-country task. Fitting and recording use question-end block outputs except for the original profile's last recorded point, which follows final normalization; the figures and Appendix~\ref{app:narrow-details} mark this exception. All interventions remain at block outputs.

The country split separates direction fitting, layer screening, and evaluation on unseen countries. Each of four disjoint groups contains six countries arranged in three cross-continent pairs. The three pairs under both requests give six questions per group: one group fits, one screens, and validation A and B test the frozen choice. The auxiliary marker questions provide the selection contrast but do not supply factual answers to any continent question.

To measure what transfers beyond fitted content and the continent task's own request contrast, we remove their overlap with the raw marker contrast. Let $\vect{\mu}_{\mathrm{first}}^{\ell,\mathrm{main}}$ and $\vect{\mu}_{\mathrm{second}}^{\ell,\mathrm{main}}$ be the averages of the three first- and three second-country vectors in the main-task fitting set. Their difference is $\vect{r}^{\ell,\mathrm{raw}}=\vect{\mu}_{\mathrm{first}}^{\ell,\mathrm{main}}-\vect{\mu}_{\mathrm{second}}^{\ell,\mathrm{main}}$. Excluding the fitted continent space gives
$\widetilde{\vect{o}}^{\ell}=(\vect{I}_{H}-\vect{P}^{\ell})\vect{o}^{\ell,\mathrm{raw}}$
and
$\widetilde{\vect{r}}^{\ell}=(\vect{I}_{H}-\vect{P}^{\ell})\vect{r}^{\ell,\mathrm{raw}}$.
We then remove the component of $\widetilde{\vect{o}}^{\ell}$ parallel to $\widetilde{\vect{r}}^{\ell}$:
\begin{equation}
\vect{o}^{\ell}=\mathtt{Unit}\!\left[
\widetilde{\vect{o}}^{\ell}
-\bigl(\mathtt{Unit}(\widetilde{\vect{r}}^{\ell})^\top
\widetilde{\vect{o}}^{\ell}\bigr)
\mathtt{Unit}(\widetilde{\vect{r}}^{\ell})
\right].
\label{eq:object-selection-direction}
\end{equation}
The resulting unit direction $\vect{o}^{\ell}$ is the transferred selection residual. Here $\mathtt{Unit}$ normalizes a nonzero vector as defined in Section~\ref{sec:parameter-routing}; the exclusion is relative to the two fitted components, not every possible encoding of content or requests. To test its contribution, center a main-task state at $\vect{b}^{\ell}=\tfrac12(\vect{\mu}_{\mathrm{first}}^{\ell,\mathrm{main}}+\vect{\mu}_{\mathrm{second}}^{\ell,\mathrm{main}})$ and delete its observed coefficient by adding
\begin{equation}
\Delta\vect{h}_{x}^{\ell,\mathrm{sel}}
=-\bigl[(\vect{o}^{\ell})^{\top}
(\vect{h}_{x}^{\ell}-\vect{b}^{\ell})\bigr]\vect{o}^{\ell}.
\label{eq:object-selection-deletion}
\end{equation}
We compare the resulting answer-margin damage with eight Gaussian changes matched to the deletion length, using the averaging rule in Equation~\ref{eq:single-query-answer-effect}. Each split is evaluated separately. A second, non-causal diagnostic asks whether the unmodified states for first- and second-country requests lie on the expected sides of $\vect{o}^{\ell}$ after centering the pair at its midpoint. This is the pointing fraction from Equation~\ref{eq:single-query-pointing} with $\vect{o}^{\ell}$ in place of the parameter-routing candidate.

A layer first has to work in the task that defines the marker contrast: unmodified marker accuracy must be at least 90\%; moving toward the opposite request must switch at least two-thirds of answers; and the strongest matched random control may switch at most one-third. The transferred residual is called stable only if its deletion effect is also positive in the screening split and in both held-out validation splits. These gates test a fitted interface across tasks; they do not establish a native, shared selection module.

\subsection{Transferred Selection Is Stable in One Model--Format Setting}

The transfer test asks whether a direction calibrated on supplied markers also influences answers that require stored facts. Across the evaluated models and formats, only Qwen adjective answers show a selection residual that passes both calibration and the separate data-split checks.

\begin{figure}[t]
\centering
\includegraphics[width=\linewidth]{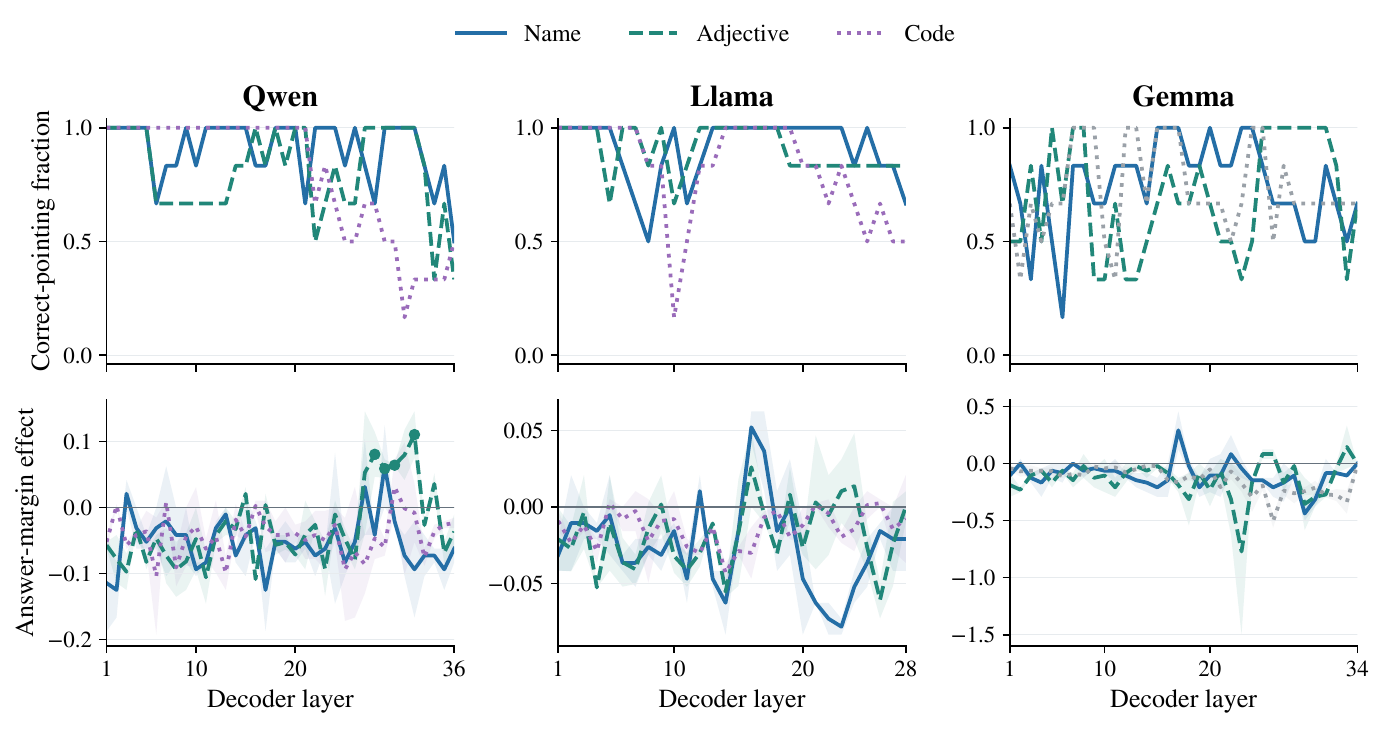}
\caption{\textbf{Transferred selection has a task-dependent layerwise effect.} Top: the paired-request correct-pointing fraction of the transferred selection direction. Bottom: its deletion effect on the answer margin relative to eight matched random changes. Lines average the two validation partitions; shading spans their values. Filled dots identify layers passing both auxiliary calibration and positive-effect checks in all three evaluated partitions. Other values are retained, including negative values. Gray dotted Gemma code curves are diagnostic because baseline qualification failed. These are effects after transfer to continent questions, not the marker-calibration switching rates.}
\label{fig:object-selection-lifecycle}
\end{figure}

Figure~\ref{fig:object-selection-lifecycle} reports all three answer formats for each model. Only Qwen's adjective task passes every calibration and split-level gate, at layers 28--30 and 32; its largest validation-mean effect is 0.11 answer-margin units. No other behaviorally qualified model--format combination passes the same criterion. Their curves remain measurements, but isolated positive values are not promoted to stable transfer.

The two results have different meanings. Marker switching shows that the raw contrast can redirect a choice when both answers are supplied. The continent result asks whether a residual of that contrast still matters after transfer to a task whose answers must come from the model. Success in one Qwen format provides a conditional contribution, not evidence for a selection stage shared across models or knowledge tasks. Failure elsewhere is likewise a boundary of this linear interface, not proof that those models never select objects.

\subsection{Direct Selection in a Separate Base-Model Task}

Transfer alone does not show that two internally represented facts remain usable after a selection change. We test that stronger property directly with Qwen2.5-3B base weights and country--capital questions. A prompt names France and Germany, for example, and requests one capital without supplying Paris or Berlin. This is a separate task and checkpoint from the three-instruction-model continent comparison.

Forty-four country pairs pass the answer gate. Fifteen pairs fit the measurements, eight select a source layer, and disjoint groups of ten and eleven pairs validate the fixed source. Two phrasings and both requested objects give 40 and 44 validation questions. Appendix~\ref{app:base-capital-selection} specifies the full fitting and restoration protocol.

We use four \emph{linear readouts} to track what changes after the edit. A readout is a fitted map from a recorded hidden-state vector to a target vector; it is an analysis tool, not a component the model runs to answer, following the use of independently fitted probes to inspect intermediate representations \citep{alain2016probes}. Write its prediction as $\widehat{\vect{y}}_{x}^{\ell,c}=(\vect{W}^{\ell,c,\mathrm{eff}})^\top\vect{h}_{x}^{\ell}+\vect{\beta}^{\ell,c,\mathrm{eff}}$, where $c$ names the target, $\vect{W}^{\ell,c,\mathrm{eff}}\in\mathbb{R}^{H\times d_c}$ and $\vect{\beta}^{\ell,c,\mathrm{eff}}\in\mathbb{R}^{d_c}$ are fitted coefficients, and $d_c$ is the target dimension. The superscript $\mathrm{eff}$ marks coefficients after absorbing fixed linear preprocessing; the explicitly standardized fit used for the instruction-model tests is defined in Appendix~\ref{app:direct-selection}.

Three targets describe the two countries and their two candidate capitals, regardless of which country is requested. We represent each name by the mean of its input-token embeddings, the numerical representations assigned before the decoder blocks. Principal-component analysis \citep{shlens2014pca}, fitted only on the fitting names, supplies compact coordinates along their main directions of variation. The first target concatenates the two country coordinates; the other two separately give the first and second capital coordinates. The fourth target is a two-entry selection label, $(1,0)^\top$ for the first object and $(0,1)^\top$ for the second. The country and capital targets stay unchanged across opposite requests; only the selection target changes.

Screening fixes decoder layer 26 at the question-end position. Replacing only the fitted selection component with its opposite-request value changes downstream selection and the answer. All three non-selection readouts remain within matched-random change on every validation question. At layers 27--30, the directed selection change exceeds eight random controls and a wrong-position control; the answer change exceeds every control on 37/40 questions in validation A and 44/44 in validation B.

Linear readout preservation still does not show that both capitals can drive an answer. The experiment therefore copies the same source-modified state into two continuations. At a receiving layer, one continuation changes only the fitted selection subspace toward the clean state for the original request; the other changes the same subspace toward the clean state for the paired opposite request. All coordinates outside that fitted subspace are left as they were in the source-modified computation. A question passes only if the two branches produce the corresponding capital answers, both changes beat random and wrong-position restorations of the same length, and the two candidate-capital readouts remain within random bounds.

At receiving layers 27, 28, and 29, the joint pass counts in validation A are 25/40, 28/40, and 24/40; in validation B they are 34/44, 34/44, and 32/44. Thus both groups meet the prespecified 60\% threshold at the same three layers. These fractions count questions for which both continuations pass the full test; they do not estimate what fraction of a hidden state is ``capital knowledge.''

An early control shows why the two-branch test is necessary. At the token position of the ordinal word (the request's first or second), an internal edit in the first decoder block---recorded as zero-based layer 0---changes later selection and answers, yet restoring the original choice at layers 15--19 recovers the original answer on at most 2/40 and 1/44 questions. This early internal edit therefore changes the answer path before two usable candidates have been demonstrated. The base-model result supports direct selection only for the layer-26-to-29 capital setting; it does not identify a universal earliest selection layer.

\subsection{Direct Selection in the Instruction-Model Continent Task}
\label{sec:direct-continent-selection}

We apply the same functional question to the Qwen instruction model and continent names. For a prompt containing Angola and Japan, can one state change redirect downstream selection and the answer while preserving readouts of both country identities and both candidate continents? The original six fit countries form 24 ordered cross-continent pairs and 48 first- or second-object questions. Four readouts separately predict the two identities, the first and second candidate continents, and the selected position. The candidate-continent labels do not change when the request changes. Appendix~\ref{app:direct-selection} gives the fitting objective, controls, and gates.

Let $\vect{o}^{\ell,\mathrm{direct}}$ be the unit direction of the scalar selection readout, fitted with target $+1$ for the first object and $-1$ for the second. For question $x$, $\vect{h}_{x}^{\ell,\mathrm{opp}}$ is the unmodified state of the same country pair and wording when the other object is requested. We replace only the coordinate of $\vect{h}_{x}^{\ell}$ along the fitted selection direction by adding
\begin{equation}
\Delta\vect{h}_{x}^{\ell,\mathrm{swap}}
=\vect{o}^{\ell,\mathrm{direct}}
(\vect{o}^{\ell,\mathrm{direct}})^\top
(\vect{h}_{x}^{\ell,\mathrm{opp}}-\vect{h}_{x}^{\ell}).
\label{eq:direct-selection-swap}
\end{equation}
The operation does not copy the full opposite-request state. It changes one fitted scalar coordinate, using a direction learned directly from the continent task rather than transferred from the marker task.

All country pairs first pass the unmodified three-continent answer comparison. Screening changes first-object requests to second-object requests only, giving 12 questions: three pairs, two country orders, and two held-out phrasings. The opposite requests supply the replacement coordinate, rather than a second set of screening edits. At each source layer 1--33, the swap is compared with two random directions formed from Gaussian combinations of the centered fitting states and the same swap applied at the word ``continent.'' A source must preserve all three non-selection readouts on at least 70\% of questions. It must also yield at least three later layers where at least 60\% of questions jointly preserve those readouts, move the selection score beyond every control, and change the answer margin beyond every control. Equation~\ref{eq:direct-answer-effect} first subtracts the strongest control within each question and then averages the 12 differences.

\begin{table}[htbp]
\centering
\caption{Direct-selection diagnostics in the Qwen instruction-model continent task. The denominator is 12 selection questions, not 12 independent country pairs. The last column is the best single receiving-layer joint fraction for that source. At least three receiving layers must each reach 60\%; with 12 questions this requires at least eight passes per layer. Rows illustrate distinct failures and are not validation-selected operating points.}
\label{tab:direct-selection-boundary}
\small
\begin{tabular}{rccr}
\toprule
Source & Source-content preservation & Answer effect & Best later joint fraction\\
\midrule
24 & $11/12$ & $0.098$ & $5/12$\\
25 & $11/12$ & $0.056$ & $6/12$\\
27 & $1/12$ & $1.034$ & $1/12$\\
\bottomrule
\end{tabular}
\end{table}

No source passes. Source 25 reaches the largest single-receiver count, 6/12, but the rule requires at least 8/12 at each of three receivers. Source 27 produces a larger mean answer effect, 1.034, while preserving the other readouts on only 1/12 questions. That combination is consistent with a broad disruptive edit, not selective control. A fresh model load reproduces every scan value and all numerical-write checks pass. The prespecified stop rule therefore blocks both held-out intervention and two-candidate restoration.

\subsection{Capital Questions across the Three Instruction Models}
\label{sec:cross-model-capital-selection}

The base-model capital result leaves a specific generality question: does the same direct-selection test work with instruction-tuned weights? We retain the capital relation and test the three instruction models used in the continent experiments. The task has 196 questions from 49 country pairs, two phrasings, and both requested objects. Qwen answers 195 correctly; Llama and Gemma answer all 196.

Fifteen pairs fit the readouts. Seven qualified Qwen pairs and eight pairs in each other model screen source layers; two disjoint groups of 13 pairs provide validation. As in the direct instruction-model continent test, selection is fitted with scalar labels $+1$ and $-1$, whereas the three non-selection targets track both country identities and both capitals. This scalar interface is distinct from the base-model experiment's two-entry selection interface. Appendix~\ref{app:cross-model-capital-selection} gives the controls and fixed acceptance rules; primary and fresh-load records agree exactly.

The outcomes separate three failure points. Qwen never identifies a source layer: its strongest content-preserving candidate changes the fitted selection score but not the answer beyond pointwise controls. Llama shows a narrow screening effect from source layer 14, with joint pass counts of 14/16 at layer 15 and 11/16 at layer 16, but only 9/16 at layer 17. Because the rule requires three receiving layers at or above 60\%, validation remains closed.

Gemma passes screening at source layer 17 and fixes receiving layers 18--20. In each 52-question held-out group, the downstream selection change exceeds every control on all 52 questions and source content is preserved on all 52. The answer change is less stable: it exceeds the strongest pointwise control on 32/52 questions in validation A and 31/52 in validation B. The resulting three-layer joint fractions are 32/52, 32/52, and 31/52 in validation A, but 31/52, 31/52, and 29/52 in validation B. Since all three receivers must reach 60\% in both groups, the experiment stops before two-candidate restoration. This is evidence for a reproducible internal selection candidate in Gemma, not a strict two-usable-candidate result.

\subsection{Selection Evidence Is Hierarchical and Task-Dependent}

The evidence forms a hierarchy rather than one universal stage. The marker task demonstrates controllable choice when both answers are supplied. Cross-task transfer produces a stable residual effect only for Qwen adjective answers. The separate Qwen base-model capital experiment satisfies the stronger same-state, two-continuation test. The instruction-model continent and capital experiments stop earlier: Gemma supplies the clearest held-out internal selection candidate, but no instruction model reaches two-candidate restoration.

These failures constrain the measured interface and task conditions, not the existence of selection somewhere in the models. Linear readouts capture only selected aspects of identity and candidate knowledge, and the tested direction need not exhaust the model's control representation. We therefore treat object selection as a conditional functional result, not a processing stage shared across models, relations, or fixed layer numbers.

\section{The Development and Use of Knowledge Content}
\label{sec:knowledge-state}
\label{sec:formation}

Does the fact content that becomes readable inside the model also support the answer, and can it support more than one expression of that answer? For Kenya, this content is the association with Africa, rather than the country name or the request to identify a continent. Using the paired-country tasks, we measure this content, delete it, and transfer a fixed content change between answer formats. The tests show persistent late answer dependence and transfer from continent names to adjectives, with arbitrary output codes providing a boundary.

\subsection{Task and Content Intervention}

To separate the factual association from its required expression, we keep the paired-country question fixed and vary the answer format as in Section~\ref{sec:object-selection}. A Kenya request can require the noun \emph{Africa}, the adjective \emph{African}, or an arbitrary code such as \emph{dax}. The code prompt supplies only the output map---for example, Africa = dax, Asia = wug, and Europe = blicket. It never states which continent contains Kenya. The factual association is therefore fixed while the required expression changes.

For each model, layer, and answer format, we fit a separate continent-content space from the six fitting questions in Section~\ref{sec:object-selection}. Denote this question set by $\mathcal{X}_{\mathrm{fit}}$ and its two questions requesting continent $k$ by $\mathcal{X}_{\mathrm{fit},k}$. Each element contributes its recorded question-end vector $\vect{h}_{x}^{\ell}$. The means are
\[
\vect{\mu}^{\ell}=\frac16\sum_{x\in\mathcal{X}_{\mathrm{fit}}}\vect{h}_{x}^{\ell},
\qquad
\vect{\mu}_{k}^{\ell}=\frac12\sum_{x\in\mathcal{X}_{\mathrm{fit},k}}\vect{h}_{x}^{\ell}.
\]
Thus grouping follows the requested country's true continent, not its first or second position. The centered group means define a two-dimensional orthonormal basis $\vect{U}^{\ell}$ by Equation~\ref{eq:single-query-knowledge-fit}. Reusing that construction does not reuse the single-country fit: these means and axes come from the paired task and required answer format. The projected state and continent references are
\begin{equation}
\vect{z}_{x}^{\ell}=\vect{P}^{\ell}
(\vect{h}_{x}^{\ell}-\vect{\mu}^{\ell}),
\qquad
\vect{\kappa}_{k}^{\ell}=\vect{P}^{\ell}
(\vect{\mu}_{k}^{\ell}-\vect{\mu}^{\ell}),
\quad
\vect{P}^{\ell}=\vect{U}^{\ell}(\vect{U}^{\ell})^{\top}.
\label{eq:content-state}
\end{equation}
For question $x$, let $k_x^+$ and $k_x^-$ be the requested and paired countries' continents, respectively. The knowledge score $\gamma_{x}^{\ell}$ is the squared distance to the paired country's continent reference minus the squared distance to the requested country's reference, as defined in Equation~\ref{eq:single-query-knowledge-score}. A positive score means that the fitted two-dimensional state is closer to the correct reference. The correct-pointing fraction is
\begin{equation}
F^{\ell,\mathrm{know}}
=\frac1N\sum_{x\in\mathcal{X}}\mathtt{Ind}(\gamma_{x}^{\ell}>0),
\label{eq:content-pointing}
\end{equation}
where $\mathcal{X}$ contains the $N$ questions being evaluated and $\mathtt{Ind}$ is the indicator defined in Section~\ref{sec:parameter-routing}.

Readability does not show that the answer uses this projection. We therefore remove the question's entire fitted content component by adding
\begin{equation}
\Delta\vect{h}_{x}^{\ell,\mathrm{know}}=-\vect{z}_{x}^{\ell}.
\label{eq:content-deletion}
\end{equation}
The fixed model then continues from that layer. We compare the resulting answer-margin damage with eight random vectors matched to the actual deletion length, using the answer effect in Equation~\ref{eq:single-query-answer-effect}, with $c=\mathrm{know}$ denoting the content deletion. Thus, pointing asks what the linear interface can read, whereas deletion asks whether the subsequent answer is more sensitive to removing that fitted component than to the tested random changes.

\subsection{Fitted Content Becomes Consequential and Persists Late}

The fitted continent content contributes to answers in all three models, and its contribution persists through late layers. We establish this by comparing content deletion with matched random changes separately in screening, validation A, and validation B, then compare the paired-task result with the natural-question Qwen result.

\begin{figure}[t]
\centering
\includegraphics[width=\linewidth]{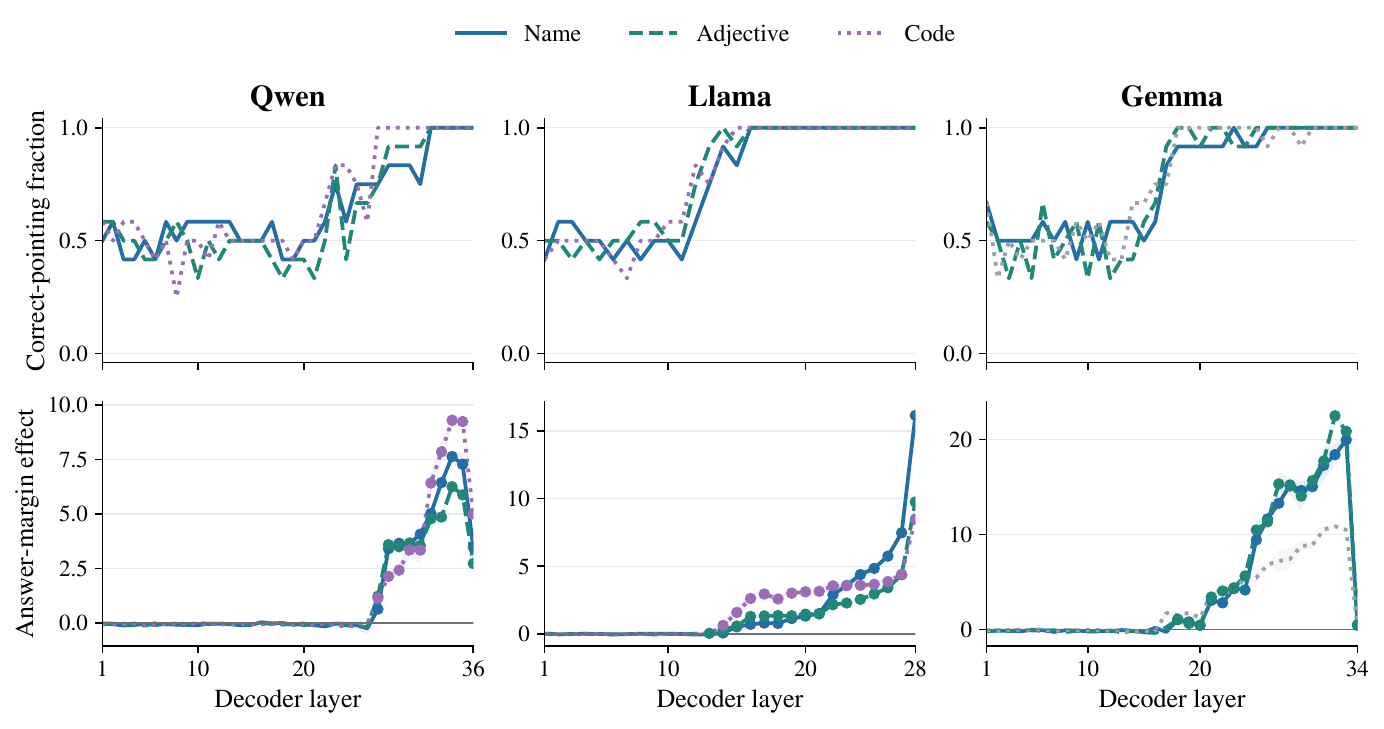}
\caption{\textbf{Answers become sensitive to deleting fitted continent content, and that sensitivity persists in late layers.} Top: the fraction of questions whose projected state is closer to the requested continent reference than to the paired alternative. Bottom: mean answer-margin damage from deleting that projection, corrected by matched random changes. Each model, layer, and answer format has its own fitted two-dimensional space. Lines and shading show the mean and range of validation A and B, not confidence intervals; filled dots require positive effects in screening and both validation splits. Gemma code results are diagnostic and shown in gray. In the original final-layer point, reading occurs after final normalization but intervention occurs before it; Appendix~\ref{app:expanded-evidence} provides location-aligned block-output profiles, and the original final drop is not interpreted as content disappearance.}
\label{fig:content-lifecycle}
\end{figure}

For continent-name answers, the screening split and both validation splits all have positive content-deletion effects at layers 27--36 in Qwen, 14--28 in Llama, and 18--34 in Gemma. Figure~\ref{fig:content-lifecycle} shows the corresponding pointing fractions and the other answer formats. The transition is not aligned by a common layer number: Qwen changes sharply near the top of its stack, while Llama and Gemma follow different profiles. Within each qualified setting, however, deleting the fitted content continues to damage the answer through the last measured layers.

The natural single-country experiment provides a separate Qwen check. Its continent space is fitted on different prompts and its interventions use float32. At layers 27--36, each pointwise 95\% interval for content-deletion answer damage lies above zero. This supports late dependence on the fitted content in that task, but the effect magnitudes are not pooled with the paired-task bfloat16 values.

\subsection{Fitted Content Transfers to Adjectives but Rarely to Codes}

The deletion result could reflect only a preference for a particular continent-name token. To test a broader role, we fit the references using noun-answer questions only. At one layer fixed before this comparison, we move an adjective or code question from the requested-continent reference toward the paired-continent reference:
\begin{equation}
\Delta\vect{h}_{x}^{\ell,\mathrm{transfer}}
=\vect{\kappa}_{k_x^-}^{\ell,\mathrm{name}}
-\vect{\kappa}_{k_x^+}^{\ell,\mathrm{name}}.
\label{eq:content-transfer}
\end{equation}
The superscript $\mathrm{name}$ marks references fitted on continent-name prompts. The target question and model weights remain fixed; no reference is refitted on adjective or code answers. For example, a Kenya question paired with China receives the noun-derived Africa-to-Asia change, and we measure whether its answer margin moves from African toward Asian.

The fixed layers are 34 for Qwen, 28 for Llama, and 33 for Gemma. For question $x$, let $m_x^0$ be the unmodified correct-minus-paired answer margin and $m_x^{\mathrm{transfer}}$ be the answer margin after the noun-derived change and $m_x^{\mathrm{random},j}$ the margin after equal-length random control $j$. For a validation partition $\mathcal{X}$, the reported transfer shift is
\[
\frac{1}{|\mathcal{X}|}\sum_{x\in\mathcal{X}}
(m_x^0-m_x^{\mathrm{transfer}})
-\max_{j=1,\ldots,8}\frac{1}{|\mathcal{X}|}\sum_{x\in\mathcal{X}}
(m_x^0-m_x^{\mathrm{random},j}).
\]
The first term is the mean loss of the correct-versus-paired answer margin after the noun-derived change; the second is the largest corresponding mean loss among eight equal-length random changes. Positive values therefore indicate a targeted shift toward the paired answer beyond every tested random direction.

A transfer passes only if this corrected shift is positive in both validation partitions and at least two of the three country pairs in each partition have positive pair means. Noun-to-adjective transfer passes in all three models. Averaging the two six-question validation shifts gives 15.72 for Qwen, 19.78 for Llama, and 46.09 for Gemma. These are answer-margin changes, not accuracies or proportions. Because the fitted reference vectors and score scales are model-specific, the three magnitudes are interpreted within model rather than ranked across models.

\begin{figure}[t]
\centering
\includegraphics[width=\linewidth]{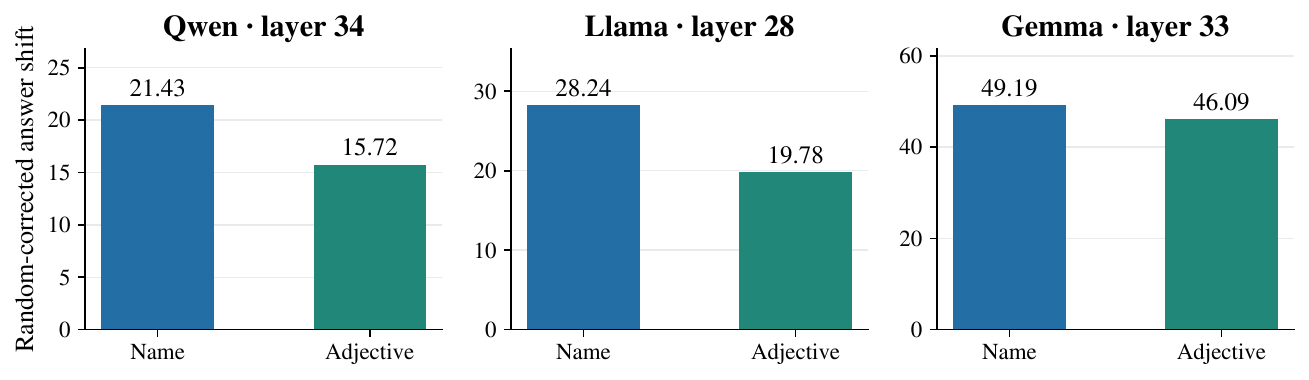}
\caption{\textbf{A fixed noun-derived knowledge change transfers to adjective answers.} Both bars use references fitted only on continent-name questions. Heights average the corrected shifts of the two validation partitions; no statistical confidence interval is asserted for these bars. Panels identify the model and the layer fixed before this test. The same direction did not generally transfer to arbitrary codes: only 1/5 baseline-qualified mappings passed in Qwen, 0/4 in Llama, and 0/2 in Gemma.}
\label{fig:content-transfer}
\end{figure}

Each model also receives six different arbitrary code maps. Before testing transfer, a map must yield the correct code on all baseline questions in both validation splits. Five maps qualify in Qwen, four in Llama, and two in Gemma. Only one Qwen map passes the transfer criterion; none pass in Llama or Gemma. The remaining maps are reported as baseline failures or transfer failures, not counted as evidence that an otherwise valid opportunity succeeded.

\subsection{Content Persists More Broadly Than Routing Effects}

Across the three models, the answer becomes sensitive to deleting the fitted continent component and remains sensitive in late layers. A noun-derived change also shifts the corresponding adjective without adjective refitting, supporting reuse across these two natural expressions. Arbitrary codes provide the boundary: most baseline-qualified maps do not transfer. The result therefore concerns a fitted component shared by noun and adjective answers, not an output-independent representation that works unchanged under every mapping.

\section{Layerwise Changes in Routing and Content Dependence}
\label{sec:joint-process}
\label{sec:handoff}
\label{sec:functional}

Does the answer become less dependent on a request component as it becomes more dependent on fact content? Sections~\ref{sec:parameter-routing}--\ref{sec:knowledge-state} establish measurements for these roles; we now compare their effects across depth under a common paired-country protocol. We test the resulting earlier-to-later contrast on new countries and check whether request specificity, fitting variation, or recovery after a single deletion explains it. Finally, we change the request-direction construction within the same task. This comparison locates the dependence shift in a particular fitted representation rather than all information about the request.

\subsection{Routing and Content Have Model-Specific Overlapping Profiles}

We first compare whether the measured request, selection, and content contributions occupy distinct or overlapping layers. For the same paired-country questions, we separately delete a request direction shared across country pairs, the transferred selection direction, and the target-content projection. Each deletion acts at the question-end position and has eight random controls matched to its own length. Holding the model, prompt, candidate answers, and evaluation questions fixed makes the three curves comparable as interventions on different fitted components.

The shared request direction is called \emph{global} because one fit is used for every evaluated country pair. The pair-conditioned direction of Section~\ref{sec:parameter-routing}, in contrast, is constructed separately for each pair. We define the global fit next before comparing its depth profile with content and selection.

At each layer, take the mean of the three first-country fitting-question vectors minus the mean of the three second-country vectors. Exclude its projection into the fitted content space and its overlap with the content-excluded marker-selection contrast. Normalizing the remainder gives $\vect{r}^{\ell,\mathrm{global}}$, with the superscript identifying a pair-shared direction. Its coefficient is measured relative to $\vect{b}^{\ell}$, the midpoint of those two fit means, and deleted by adding
\begin{equation}
\Delta\vect{h}_{x}^{\ell,\mathrm{par}}
=-\bigl[(\vect{r}^{\ell,\mathrm{global}})^\top
(\vect{h}_{x}^{\ell}-\vect{b}^{\ell})\bigr]\vect{r}^{\ell,\mathrm{global}},
\label{eq:joint-parameter-deletion}
\end{equation}
Both $\vect{r}^{\ell,\mathrm{global}}$ and $\vect{b}^{\ell}$ are fixed from the fitting questions before validation. The resulting intervention therefore tests the shared request component, whereas the natural-question intervention uses a separately constructed direction and midpoint for each pair.

The three profiles overlap rather than forming a shared sequence. The global request candidate has positive answer effects in model- and format-specific layer sets, which overlap the onset of content effects. Content deletion remains consequential later. The transferred selection residual is stable only for Qwen adjective answers, where its effect overlaps rather than follows the parameter-candidate interval. Appendix~\ref{app:narrow-details} gives the full profiles.

The complementary timing analysis in Appendix~\ref{app:route-strength-timing} compares the natural request coordinate with its answer and later-knowledge effects. Qwen shows a mid-layer coordinate rise before sustained content dependence; Gemma shows a more overlapping partial analogue; Llama lacks a sustained route-effect window under those checks. Because this analysis uses the paired task and its global direction, it complements rather than directly replicates the natural-question result.

\subsection{Global-Route Dependence Decreases on New Countries}

We test whether the earlier-to-later decline of global-route dependence persists beyond the countries used to construct and screen it. The new evaluation contains 48 countries absent from every original split, arranged as 24 non-overlapping pairs. Two country orders, two held-out wordings, and both requested objects give eight questions per pair, or 192 questions. None is used to fit a direction or content space.

To test an existing depth contrast rather than select one on the new countries, we fix the comparison layers from the original evidence. In 1-based numbering, Qwen compares $\{29,31\}$ with $\{32,34\}$, Gemma compares $\{24\}$ with $\{25,30\}$, and Llama compares $\{15,24\}$ with $\{28\}$. The earlier and later sets summarize selected regions, including nonadjacent layers; they do not define continuous processing stages.

For component $c$ and layer set $\mathcal{L}$, $E_{\mathcal{L},c}^{\mathrm{ans}}$ uses three aggregation steps. It first averages damage over the selected layers and all eight questions within each pair, then averages the 24 pair means. Finally, it subtracts the largest mean obtained by applying the same aggregation to each of eight random controls (Appendix~\ref{app:expanded-evidence}). We use two contrasts:
\begin{equation}
\begin{aligned}
\Delta E_{\mathrm{early-late}}
&=E_{\mathcal{L}_{\mathrm{early}},\mathrm{par}}^{\mathrm{ans}}
-E_{\mathcal{L}_{\mathrm{late}},\mathrm{par}}^{\mathrm{ans}},\\
\Delta E_{\mathrm{knowledge-route}}
&=E_{\mathcal{L}_{\mathrm{late}},\mathrm{know}}^{\mathrm{ans}}
-E_{\mathcal{L}_{\mathrm{late}},\mathrm{par}}^{\mathrm{ans}}.
\end{aligned}
\label{eq:handoff-contrasts}
\end{equation}
The first contrast asks whether deleting the global route causes more damage in the earlier set than in the later set. The second asks whether deleting fitted content causes more damage than deleting that route in the later set. Both contrasts use the same 24 country pairs. We resample pairs 2,000 times, keeping each pair's eight variants together, to obtain 95\% percentile intervals.

\begin{figure}[t]
\centering
\includegraphics[width=\linewidth]{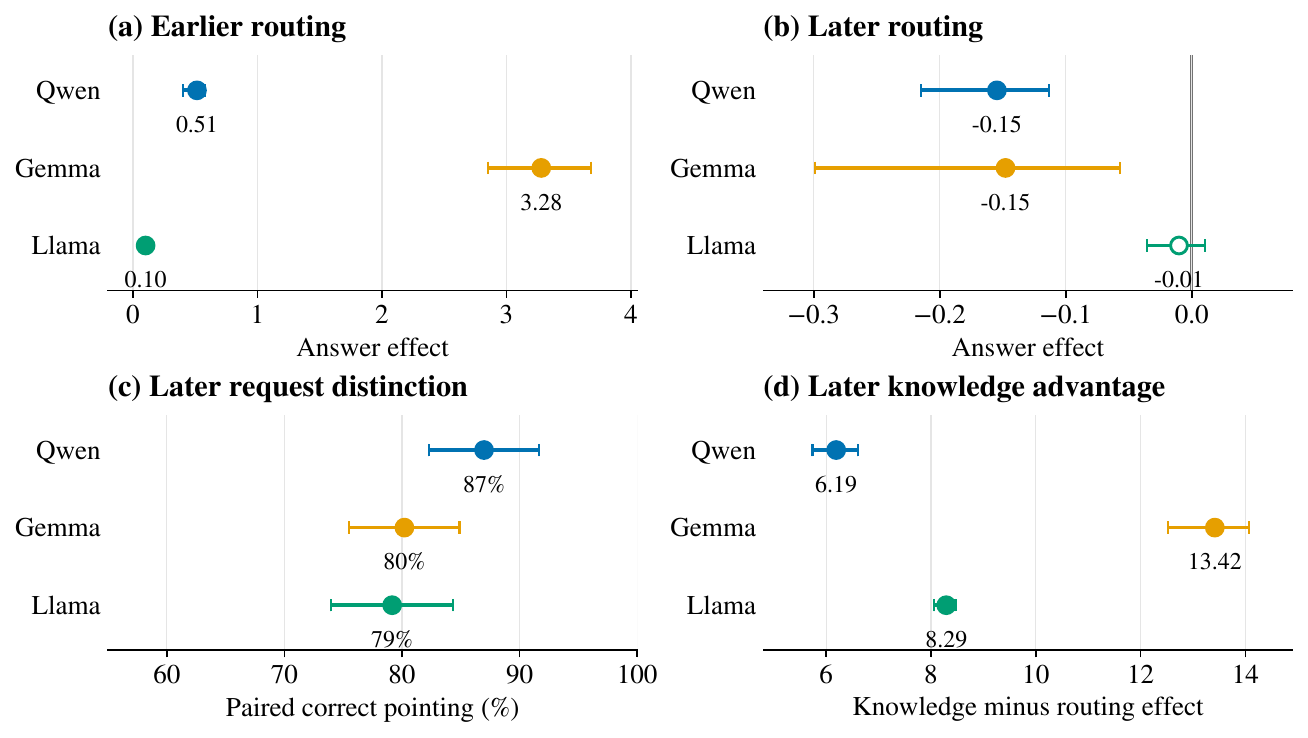}
\caption{\textbf{Deletion of the fitted global request direction causes less answer damage in fixed later layer sets, while fitted-content deletion remains consequential.} All three models use the same 24 new country pairs but model-specific layer sets fixed before this evaluation: Qwen $\{29,31\}$ versus $\{32,34\}$, Gemma $\{24\}$ versus $\{25,30\}$, and Llama $\{15,24\}$ versus $\{28\}$. Request states remain separable along the late global direction. Bars show 95\% country-pair bootstrap intervals; a hollow marker denotes an interval spanning zero. The figure concerns this global fitted direction, not all request representations or a shared processing boundary.}
\label{fig:expanded-handoff}
\end{figure}

The intervals for both registered contrasts in Equation~\ref{eq:handoff-contrasts} lie above zero in all three models (Figure~\ref{fig:expanded-handoff}); the Llama later-route point itself spans zero. Thus, the answer is less sensitive to deleting the fitted global route in the later set, yet remains more sensitive to deleting fitted content there. Llama does not meet the same fitting-robustness gate as Qwen and Gemma. The cross-model agreement is limited to the two contrast signs; separately fitted answer-margin magnitudes are not ranked across models, and the result does not imply aligned layers or the same physical implementation.

The late states remain separable along the same global direction, so its lower deletion effect is not evidence that request information has vanished. We use \emph{routing--content handoff} as shorthand for one operational observation only: relative to matched random changes, late answers are less sensitive to deleting this global request direction and remain sensitive to deleting the fitted continent content. The term does not name a discrete model module or a universal stage. The next controlled comparison tests whether another request direction has the same profile.

\subsection{Specificity and Repeated-Deletion Controls Preserve the Shift}

The declining deletion effect could reflect nonspecific damage, sensitivity to the small fitting set, or recovery immediately after the edit. We test these alternatives by examining internal knowledge effects and raw damage, varying the fit and request labels, and repeating deletion across successive layers.

Some source-route deletions reduce later fitted knowledge as well as the answer margin (Appendix~\ref{app:narrow-details}), linking the edit to an internal outcome. In Qwen and Gemma, raw targeted answer damage itself decreases, so the corrected decline is not explained solely by a rising random baseline or shrinking deletion length. Four fits---the full three-pair fit and three fits leaving out one pair---are each tested with three random seeds. Qwen and Gemma retain the earlier positive route effect in at least three fits; their later route effects fail and later content effects pass. Llama's earlier route does not meet this fitting-robustness criterion.

For Qwen, we additionally replace the true first-versus-second labels with each of nine balanced but incorrect labelings of the six fit examples. Pooled over the prespecified set $\{29,31\}$, the true direction causes more damage than the strongest wrong-label direction both at its natural deletion length and after vector-length matching. Layer 29 alone has an interval spanning zero, while layer 31 passes both comparisons (Appendix~\ref{app:expanded-evidence}). The pooled control rules out the claim that any balanced partition of the small fit set produces the same earlier-set effect.

Finally, a single deletion might be repaired by the next layer. We therefore compare deleting the global route once at layer 32 with recomputing and deleting its coefficient at every layer from 32 through 36. For each country pair, the extra raw margin damage is divided by that pair's unmodified margin and then averaged. The result is 0.21\%, with a one-sided 95\% upper bound of 0.45\%, below the prespecified 10\% practical threshold. Repeatedly removing this fitted component therefore does not recover a substantial late answer dependence.

\begin{figure}[t]
\centering
\includegraphics[width=\linewidth]{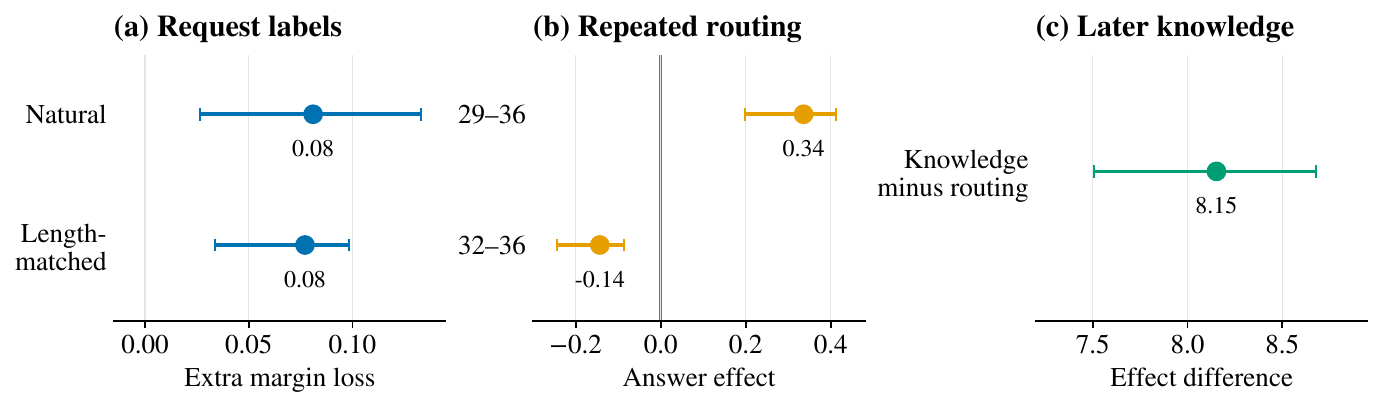}
\caption{\textbf{Qwen controls test request specificity and repeated deletion.} (a) The true request direction exceeds the strongest of nine balanced wrong-label directions, averaged over layers 29 and 31. (b) Repeated routing-deletion effects use random paths matched to the actual lengths after numerical rounding. (c) Repeated late content deletion exceeds repeated late routing deletion. Points and 95\% intervals use the same 24 country pairs. The single-versus-repeated comparison concerns the measured direction, not every possible representation of a request.}
\label{fig:expanded-controls}
\end{figure}

\subsection{Pair-Conditioned Routes Retain Late Dependence}
\label{sec:matched-direction-comparison}

Is the late decline preserved when the same paired-country task uses a pair-conditioned request direction instead of the global fit? We hold Qwen instruction weights, float32 precision, question-end block outputs, candidate answers, and the 24 validation pairs fixed, and vary direction construction and centering separately. The resulting four combinations let us distinguish an effect tied to the global deletion from one shared by both constructions.

The direction is either the fit-set contrast shared across pairs or a contrast built for the current pair from template A. The center is either the midpoint of fit-set requests or the current pair's midpoint under the evaluated template. Templates B and C evaluate every combination. Appendix~\ref{app:same-condition} specifies the corresponding input indices and fitted objects.

For a combination $c$, write $\vect{r}_{x}^{\ell,c}$ for its unit request direction and $\vect{b}_{x}^{\ell,c}$ for its center. The deletion and its intended length are
\begin{equation}
\begin{aligned}
\Delta\vect{h}_{x}^{\ell,c}
&=-\bigl[(\vect{r}_{x}^{\ell,c})^\top
(\vect{h}_{x}^{\ell}-\vect{b}_{x}^{\ell,c})\bigr]\vect{r}_{x}^{\ell,c},\\
\|\Delta\vect{h}_{x}^{\ell,c}\|_2
&=\left|(\vect{r}_{x}^{\ell,c})^\top
(\vect{h}_{x}^{\ell}-\vect{b}_{x}^{\ell,c})\right|.
\end{aligned}
\label{eq:controlled-deletion}
\end{equation}
Because $\vect{r}_{x}^{\ell,c}$ has unit length, the deletion length is the absolute projection coefficient in the second line. Changing either the direction or the center therefore changes not only which component is removed but also how much is removed. Each condition is compared with random vectors matched to its own actual length. The targeted deletions themselves are not equalized across conditions; their lengths are therefore also reported when interpreting the result.

\begin{table}[htbp]
\centering
\caption{\textbf{Late answer dependence changes with request-direction construction and natural deletion length.} Under the same Qwen paired task, deleting the global fitted direction causes less damage than its strongest matched random control, whereas deleting a pair-conditioned direction remains consequential under either center. Effects average layers 32--36 and four questions per country pair, then subtract the largest mean loss among eight actual-length-matched random controls. Intervals resample the same 24 pairs 2,000 times. All rows use float32; the natural-question row uses its own fitted knowledge space.}
\label{tab:controlled-directions}
\small
\begin{tabular}{llll}
\toprule
Task & Request direction & Center & Effect [95\% interval]\\
\midrule
Paired & Global fit & Fit & $-0.195$ [$-0.251$, $-0.160$]\\
Paired & Global fit & Pair & $-0.034$ [$-0.078$, $-0.009$]\\
Paired & Pair-conditioned & Fit & $3.092$ [$2.464$, $3.624$]\\
Paired & Pair-conditioned & Pair & $3.051$ [$2.438$, $3.581$]\\
Natural & Pair-conditioned & Pair & $2.401$ [$1.760$, $2.971$]\\
\bottomrule
\end{tabular}
\end{table}

Table~\ref{tab:controlled-directions} changes the interpretation of the earlier handoff result. Within the same paired task, pair-conditioned deletion has a positive late effect under either center, whereas global deletion causes less damage than its strongest tested random control. Changing only the center does not reproduce the difference. The mean actual deletion length at layers 32--36 is 16.38 for the global direction with the fit center and 80.47 for the pair-conditioned direction with that same center. Thus, direction construction and intervention length change together. The experiment shows dependence on the deletion definition; it does not determine whether orientation, length, or their interaction produces the larger pair-conditioned effect.

Using the pair-conditioned direction and pair center in both tasks leaves both late effects positive, with a smaller effect for natural single-country questions. Task format alone therefore cannot explain why the natural-question candidate remains consequential while the paired task's global candidate does not. Appendix~\ref{app:same-condition} reports the direct contrasts and later-knowledge measurements.

\subsection{The Joint Account Is Representation- and Model-Specific}

The joint comparison shows a change in dependence on a particular request representation, alongside persistent dependence on fitted content. Request distinctions can be readable before their deletion affects answers. Later, content deletion remains consequential while the global request deletion loses its earlier effect on new countries; repeated deletion does not restore a substantial late dependence. The matched comparison adds an essential distinction: a pair-conditioned request deletion retains a late effect in the same paired task. The depth profile therefore belongs to the specified direction, center, and intervention length, not to every representation of the request.

Selection supplies a separate functional result rather than a universal next stage in that trajectory. Marker-task control transfers stably in one instruction-model format, and the strict test of two still-usable candidates succeeds in the separate Qwen base-model capital task. The instruction-model continent and capital tests stop before restoration (Sections~\ref{sec:direct-continent-selection} and \ref{sec:cross-model-capital-selection}). Together, these findings distinguish controlling which fact is formed from selecting among available facts without imposing one processing order on all models.

\section{Limitations}
\label{sec:limitations}

The results establish functional distinctions under specific tasks and fitted measurements. Their scope is limited by the knowledge relations and models tested, the information captured by the linear measurements, and the interventions used to assess dependence. These limitations qualify how far the findings generalize without changing the within-task results.

The main evidence concerns country--continent associations. Paired-country experiments cover three instruction-tuned models, whereas the natural single-country intervention is complete only for Qwen. Nouns, adjectives, and arbitrary codes change how one association is expressed; they are not three knowledge domains. The expanded evaluation uses 24 held-out country pairs, but all fitted directions and content spaces are estimated from small country sets. These choices limit population, relation, and fitting-sample generality.

The natural-route timing analysis reuses frozen Qwen validation states after the original experiment and therefore supplies a diagnostic onset rather than a prospectively registered change point. Only Qwen has the full natural single-country states needed to divide the pair-conditioned coefficient by the whole-state scale. The three-model paired-task analysis uses a different global direction and normalizes by fitted request separation; it can reveal complementary model-specific trajectories but is not a direct Llama/Gemma replication of the Qwen timing result.

The country--continent prompts do not state the target association. Supplied markers appear only in the auxiliary selection task, and code prompts provide only a continent-to-code output map. Nevertheless, the evaluated instruction checkpoints combine pretraining with later training. The experiments study how the released weights use these associations; they do not assign each fact exclusively to pretraining rather than instruction tuning or another training stage.

Every named route and content component is a fitted linear measurement at the final input position. The two-dimensional continent space is not an exhaustive account of how a model represents the fact, and projecting it out does not remove every possible encoding. Likewise, the pair-conditioned natural-question direction is built separately for each country pair. Its pointing score tests transfer from one wording to two others within that pair; it is not a classifier trained on old countries and applied independently to unseen identities.

The common-protocol Qwen comparison holds the model, paired task, precision, state position, and evaluation pairs fixed. It shows that changing the request-direction construction changes the late-layer result. However, each deletion removes that direction's natural projection, so direction and intervention length change together: the mean late deletion lengths are 16.38 for the global candidate and 80.47 for the pair-conditioned candidate under the same fit center. We did not compare the two orientations at one common targeted length. The reported handoff therefore concerns deletion of the fitted global direction, not all request representations, and the fitted content is not shown to be its unique mediator. In the natural single-country experiment, deletion-based loss of later knowledge also remains inconclusive after pooling sixteen random controls; only reversal gives the confirmed later-knowledge effect.

Object-selection evidence is task- and checkpoint-dependent. A transferred marker-selection residual is stable only for Qwen adjective answers. The direct Qwen instruction-model continent test finds no qualifying source. In instruction-model capital tests, Qwen again finds no source, Llama stops after only two qualifying receivers, and Gemma carries a selection change to fixed held-out receivers but does not pass the three-layer answer criterion in both groups. None reaches two-candidate restoration. These failures constrain the tested directions and finite readouts; they do not show that the models lack selection. The full two-continuation criterion succeeds only in the separate Qwen base-model capital task and cannot fill the instruction-model gap. Finally, noun-derived content transfer succeeds for adjective answers but generally fails for arbitrary codes. These boundaries define the functional comparisons supported by the paper.

\section{Conclusion}

Our experiments distinguish information specifying a requested fact from content supporting its answer, and show that their contributions change across depth. This distinction requires four separate measurements: whether a request direction is readable, how strong its natural coefficient is, whether changing it affects the answer, and whether changing it affects later fitted knowledge. In natural single-country Qwen questions, the pair-conditioned distinction is readable from the first layer, its coefficient relative to the full-state scale begins a sustained mid-layer rise, and its later-knowledge effect emerges while fitted content is still becoming answer-relevant. Reversal changes both the answer and later fitted knowledge; deletion confirms answer dependence but does not pass the fixed layer-32-to-36 later-knowledge test. The result is an overlapping route-strength--content-formation trajectory, not a discrete route-complete-then-content pipeline.

The paired-country evidence shows why this trajectory must remain model- and representation-specific. Across three instruction models, dependence on a global request candidate decreases from fixed earlier to later layer sets while dependence on fitted content persists. New countries, request-specificity controls, and repeated deletion support this operational handoff. Yet a matched Qwen comparison shows that a pair-conditioned request deletion retains a late effect where the global deletion does not, and the two natural deletions differ substantially in length. Gemma provides a partially overlapping mid-layer analogue, whereas Llama has no sustained routing-effect window under the same gates. Thus the inferred depth profile depends on the fitted representation, center, intervention size, model, and task.

A noun-derived content change transfers to adjective answers without adjective refitting, but rarely to arbitrary codes. Direct selection among two still-usable facts is supported in one Qwen base-model capital setting; the instruction-model tests stop earlier, with Gemma showing a reproducible held-out selection candidate that does not meet the frozen answer gate. Together, these results provide a causal account in which query controls and answer-supporting content make distinct, changing contributions across depth. The account is strongest when each claim names the fitted representation, intervention, control, task, and checkpoint rather than imposing one common processing order on all models.

\subsection*{AI use statement}

In this work, we used generative AI tools to propose and refine hypotheses, provide feedback on
experimental methodology, assist with experiment implementation and result interpretation,
discover and summarize relevant literature, prepare figures and tables, and draft and edit the
manuscript. We did not treat AI-generated text as an independent source of empirical evidence or
bibliographic authority: numerical claims were checked against stored experiment outputs, code was
executed and tested, and cited sources were inspected for support of the associated claims. We
reviewed all AI-assisted work and take responsibility for the final content of this paper, including
text, claims, code, analyses, and artifacts produced with the aid of generative AI.

\subsection*{Reproducibility statement}

Each result section states the question, fitted measurement object, intervention, control, score, and decision rule. Appendix~\ref{sec:setup} defines the shared paired-country protocol. Appendices~\ref{app:narrow-details}--\ref{app:same-condition} provide the remaining fitting, centering, calibration, repeated-intervention, direct-selection, and robustness details. Available plotting data, file hashes, and plotting programs accompany the source under \path{anc/figures/}.

Frozen records retain model and tokenizer identities, rendered prompts, intervention positions, per-question outputs, and fresh-load comparisons. Compact evidence for the base-model and instruction-model capital-selection experiments is included under \path{anc/evidence/object_selection_capital/}; the natural-route timing and complementary three-model reanalyses are under \path{anc/evidence/route_strength_timing/}. Both directories include source hashes. Fresh loads check computational reproducibility and are never counted as additional statistical samples.

The original paired-task bands and the complementary three-model trajectory figure show the range of two validation splits. Country-pair bootstrap intervals are used in the natural single-country, expanded-country, and common-protocol Qwen analyses; the direct capital-selection experiment reports fixed split-level question fractions. The natural-question, common-protocol Qwen, and direct capital-selection experiments use float32 and verify the vector actually written. Whenever a screening gate fails, the paper states that held-out intervention or two-candidate restoration was not run rather than treating the absent records as missing data.

\bibliographystyle{plainnat}
\bibliography{references}
\appendix
\section{Paired-Question Protocol Details}
\label{sec:setup}

This appendix collects the common construction and scoring details of the paired-country experiments in Sections~\ref{sec:object-selection}--\ref{sec:joint-process}. The natural single-country experiment has its own pair-conditioned construction in Section~\ref{sec:parameter-routing}. The formulas below specify the paired task: first its questions and fitting data, then the three measured components, the interventions, and the outcome scores. This keeps the paired-task construction separate from the natural single-country definition.

\subsection{Task and Data}

\paragraph{A question with two possible knowledge targets.}
Each input lists two countries from different continents and asks about one of them. For example:
\begin{quote}
\small
First country: Kenya\\
Second country: China\\
Which continent is the \textbf{first} country located in?\\
Answer with only the continent name.
\end{quote}
The paired question changes only \emph{first} to \emph{second}. Its answer changes from Africa to Asia. The input supplies country names and the request, but neither country--continent association. The model must therefore use its stored knowledge to answer. Choosing countries from different continents ensures that changing the request also changes the target knowledge.

\paragraph{One association, three answer formats.}
We vary how the same knowledge must be expressed: a continent name, its adjective, or an arbitrary code. The corresponding answers in this example are:
\begin{center}
\small
\begin{tabular}{lll}
\toprule
Answer format & Request Kenya & Request China\\
\midrule
Continent name & Africa & Asia\\
Adjective & African & Asian\\
Code & dax & wug\\
\bottomrule
\end{tabular}
\end{center}
The code input provides the mapping Africa = dax, Asia = wug, Europe = blicket. It does not say which continent Kenya or China belongs to. These formats test whether the layerwise pattern depends on how a country--continent association is expressed; they are not three different knowledge domains.

\paragraph{Fitting and held-out evaluation.}
The original data contain 24 countries from Africa, Asia, and Europe. We divide them into four country-disjoint partitions. Each partition has six countries arranged into three pairs, giving six questions per answer format when both requests are included. Each continent occurs once in the first position and once in the second position. Thus, the requested position is not tied to one continent.

The \emph{fit partition} supplies the hidden states used to construct the measured directions and knowledge space; this does not train or edit model parameters. The \emph{selection partition} screens their effects. \emph{Validation A} and \emph{validation B} test whether those effects recur on new countries. The separate marker task described below supplies only the selection calibration, not the main knowledge answers.

We also evaluate 48 additional countries that occur in none of these four partitions. They form 24 non-overlapping pairs. Two country orders, two question phrasings, and two requests give eight variants per pair, or 192 questions per model. The components are constructed using only the original fit questions, never the new countries. This tests transfer across countries, wording, and order.

\paragraph{Models and baseline answers.}
We use Qwen-2.5-3B-Instruct \citep{qwen2024qwen25}, Llama-3.2-3B-Instruct \citep{dubey2024llama3}, and Gemma-3-4B-Instruct \citep{gemmateam2025gemma3}. We compare the probabilities of the requested and alternative answers, using the sequence score defined below. A model--format combination enters the original main comparison only if it prefers the correct answer on every question in selection and both validation partitions. Gemma's code format reaches five of six in one partition and is reported as diagnostic. All three models prefer the correct candidate on all 192 expanded questions.

\subsection{Three Measured Parts of a Hidden State}
\label{sec:representations}

The three components measure answer-supporting content and two candidate controls within the same recorded state. We first fix this state and its task-specific coordinates, then define the content projection and the two request contrasts.

Let $\vect{h}_{x}^{\ell}\in\mathbb{R}^{H}$ be the recorded hidden-state vector for question $x$ at layer $\ell$. Here, $H$ is the model's hidden-state dimension and $\ell\in\{1,\ldots,L\}$ indexes its $L$ decoder layers. We use the final input-token position after the model's chat template, immediately before the answer continuation, and call it the \emph{question-end position}. The same position is used across layers. Extraction details, including the final normalization, are given in Appendix~\ref{app:narrow-details}.

\paragraph{Target-knowledge state.}
For a request about Kenya, the target knowledge is Kenya's continent, Africa; for a request about China, it is Asia. We fit a two-dimensional space from the mean states for Africa, Asia, and Europe. Its orthonormal basis is $\vect{U}^{\ell}\in\mathbb{R}^{H\times2}$, and $\vect{P}^{\ell}=\vect{U}^{\ell}(\vect{U}^{\ell})^{\top}$ projects a state onto that space. We denote the six main-task fitting questions by $\mathcal{X}_{\mathrm{fit}}$ and their two questions requesting continent $k$ by $\mathcal{X}_{\mathrm{fit},k}$. Their vector means are $\vect{\mu}^{\ell}=\tfrac16\sum_{x\in\mathcal{X}_{\mathrm{fit}}}\vect{h}_{x}^{\ell}$ and $\vect{\mu}_{k}^{\ell}=\tfrac12\sum_{x\in\mathcal{X}_{\mathrm{fit},k}}\vect{h}_{x}^{\ell}$. Each model, layer, and answer format has its own fitted space.

\paragraph{Parameter-retrieval routing.}
We construct a candidate from the mean state for first-country requests minus the mean state for second-country requests. This contrast changes which stored association is requested. We test whether a direction outside the fitted knowledge space influences the answer and the target-knowledge state in later layers. We use \emph{parameter routing} for this measured direction, as in the Introduction.

\paragraph{Object-selection routing.}
The hidden-state-routing candidate comes from an auxiliary input that supplies arbitrary markers, such as Kenya = dax and China = fep. Its question asks for the marker of the first or second record. Because the markers are already supplied, this task calibrates selection of represented information. We fit its first-minus-second request direction and require a shift toward the other request to switch the preferred marker more reliably than random changes. We then transfer the direction to the continent task. We call this candidate \emph{object-selection routing}, or \emph{hidden-state routing}.

For each routing candidate, we remove its knowledge-space projection and then its projection onto the other knowledge-excluded candidate. The superscript $\mathrm{global}$ identifies a direction shared by all country pairs within this model and answer format. The resulting unit directions are $\vect{r}^{\ell,\mathrm{global}}$ for parameter routing and $\vect{o}^{\ell}$ for object selection. We call them the candidates' \emph{unique parts}. Their full construction is given in Appendix~\ref{app:fitting-formulas}. Deleting either tests a contribution outside the fitted knowledge space.

\subsection{Component Deletion and Matched Random Changes}
\label{sec:deletion}

We delete one component while leaving model parameters and all other token positions unchanged. Let $\vect{b}^{\ell}$ be the midpoint of the two main-task fit means for first- and second-country requests. The state changes computed for the three components are
\begin{equation}
\begin{aligned}
\Delta\vect{h}_{x}^{\ell,\mathrm{par}}
&=-\bigl[(\vect{r}^{\ell,\mathrm{global}})^{\top}(\vect{h}_{x}^{\ell}-\vect{b}^{\ell})\bigr]\vect{r}^{\ell,\mathrm{global}},\\
\Delta\vect{h}_{x}^{\ell,\mathrm{sel}}
&=-\bigl[(\vect{o}^{\ell})^{\top}(\vect{h}_{x}^{\ell}-\vect{b}^{\ell})\bigr]\vect{o}^{\ell},\\
\Delta\vect{h}_{x}^{\ell,\mathrm{know}}
&=-\vect{P}^{\ell}(\vect{h}_{x}^{\ell}-\vect{\mu}^{\ell}).
\end{aligned}
\label{eq:deletions}
\end{equation}
The superscripts $\mathrm{par}$, $\mathrm{sel}$, and $\mathrm{know}$ identify parameter routing, object selection, and target knowledge. We add the chosen vector to the specified decoder-layer output at the question-end position, then continue computation. A standard layerwise test intervenes at one layer only. The repeated-deletion experiment in Section~\ref{sec:handoff} instead applies the operation at each selected layer. Neither protocol intervenes at later answer-token positions.

An arbitrary change can also damage an answer. For each component $c\in\{\mathrm{par},\mathrm{sel},\mathrm{know}\}$, we therefore generate eight independent Gaussian vectors $\vect{\epsilon}_{x}^{\ell,c,j}\sim\mathcal{N}(\vect{0},\vect{I}_{H})$, with random-control index $j\in\{1,\ldots,8\}$. Here, $\vect{I}_{H}$ is the $H$-dimensional identity matrix. We scale each random change to the deletion length:
\begin{equation}
\Delta\vect{h}_{x}^{\ell,c,j}
=
\left\|\Delta\vect{h}_{x}^{\ell,c}\right\|_2
\frac{\vect{\epsilon}_{x}^{\ell,c,j}}
{\left\|\vect{\epsilon}_{x}^{\ell,c,j}\right\|_2}.
\label{eq:random-match}
\end{equation}
Thus, the comparison changes direction while matching the size of the perturbation. It does not replace the entire hidden state with a random state or copy a component from another question. Appendix~\ref{app:expanded-evidence} describes numerical-precision matching for repeated interventions.

\subsection{Information Present and Effects on Computation}
\label{sec:scores}

\paragraph{Candidate answers and deletion damage.}
For each question $x$, the candidate set $\mathcal{A}_{x}=\{a_x^+,a_x^-\}$ contains the requested answer and the other country's answer in the required format. Let $a=(a_1,\ldots,a_{T_a})$ be a candidate with $T_a$ tokens. Under intervention condition $e$, its sequence score $\lambda_{x,a}^{e}$ and the answer margin $m_x^e$ are
\begin{equation}
\begin{aligned}
\lambda_{x,a}^{e}
&=\frac{1}{T_a}\sum_{t=1}^{T_a}
\log\mathtt{Prob}(a_t\mid x,a_{<t},e;\theta),\\
m_x^{e}&=\lambda_{x,a_x^+}^{e}-\lambda_{x,a_x^-}^{e},
\qquad
\Delta m_x^{e}=m_x^0-m_x^{e}.
\end{aligned}
\label{eq:answer-margin}
\end{equation}
$\mathtt{Prob}$ is the model's next-token probability under its fixed parameters $\theta$; $a_{<t}$ is that candidate's preceding tokens. Condition $e=0$ means no intervention. A positive margin means the correct candidate receives the higher score. $\Delta m_x^e$ is the loss in that margin. This is a controlled two-candidate comparison, not unrestricted generation accuracy.

\paragraph{Answer effect beyond random changes.}
For $N=|\mathcal{X}|$ questions in an evaluation set $\mathcal{X}$, write $\Delta m_{x}^{\ell,c}$ for true-deletion damage and $\Delta m_{x}^{\ell,c,j}$ for damage from random control $j$, both computed by Equation~\ref{eq:answer-margin}. The answer effect is
\begin{equation}
E_{c}^{\ell,\mathrm{ans}}
=
\frac{1}{N}\sum_{x\in\mathcal{X}}\Delta m_{x}^{\ell,c}
-
\max_{j=1,\ldots,8}
\frac{1}{N}\sum_{x\in\mathcal{X}}\Delta m_{x}^{\ell,c,j}.
\label{eq:answer-effect}
\end{equation}
The superscript $\mathrm{ans}$ denotes an answer effect. A positive value means that deleting the fitted component causes greater average damage than every tested random direction. Negative values mean less damage than the strongest random control, not necessarily no raw deletion damage. We average over questions before taking the maximum over controls.

\paragraph{Target knowledge in later states.}
Let $\vect{z}_{x}^{\ell}=\vect{P}^{\ell}(\vect{h}_{x}^{\ell}-\vect{\mu}^{\ell})$ be the projected, centered state. For each continent $k$ in $\mathcal{K}=\{\mathrm{Africa},\mathrm{Asia},\mathrm{Europe}\}$, $\vect{\kappa}_{k}^{\ell}$ is its projected fit-mean reference. For question $x$, let $k_x^+$ and $k_x^-$ denote the requested and alternative continents. The knowledge score is
\begin{equation}
\gamma_{x}^{\ell}
=
\left\|\vect{z}_{x}^{\ell}-\vect{\kappa}_{k_x^-}^{\ell}\right\|_2^2
-
\left\|\vect{z}_{x}^{\ell}-\vect{\kappa}_{k_x^+}^{\ell}\right\|_2^2.
\label{eq:knowledge-score}
\end{equation}
Positive values mean the state is closer to the requested continent's reference. This score concerns the internal knowledge representation, whereas Equation~\ref{eq:answer-margin} concerns answer-token probabilities.

After intervening at layer $\ell$, we measure knowledge-score losses at each later layer $\ell'>\ell$. Let $\Delta\gamma_{x}^{\ell\rightarrow\ell',c}$ be the unmodified score minus the score after true deletion, and let $\Delta\gamma_{x}^{\ell\rightarrow\ell',c,j}$ be the analogous random-control loss. The later-knowledge effect is
\begin{equation}
E_{c}^{\ell\rightarrow\ell',\mathrm{know}}
=
\frac{1}{N}\sum_{x\in\mathcal{X}}\Delta\gamma_{x}^{\ell\rightarrow\ell',c}
-
\max_{j=1,\ldots,8}
\frac{1}{N}\sum_{x\in\mathcal{X}}\Delta\gamma_{x}^{\ell\rightarrow\ell',c,j}.
\label{eq:later-knowledge-effect}
\end{equation}
This tests whether the intervention changes later target knowledge as well as the final answer.

\paragraph{Correct pointing without intervention.}
We separately measure whether a component distinguishes the correct target in unmodified states. For routing, the reference $\vect{b}_{x}^{\ell,\mathrm{pair}}$ is the midpoint of the two states for the same country pair under opposite requests. Set $\sigma_x=+1$ for a first-country request and $-1$ for a second-country request. The signed routing values are $\alpha_{x}^{\ell,\mathrm{par}}=\sigma_x(\vect{r}^{\ell,\mathrm{global}})^{\top}(\vect{h}_{x}^{\ell}-\vect{b}_{x}^{\ell,\mathrm{pair}})$ and $\alpha_{x}^{\ell,\mathrm{sel}}=\sigma_x(\vect{o}^{\ell})^{\top}(\vect{h}_{x}^{\ell}-\vect{b}_{x}^{\ell,\mathrm{pair}})$. The correct-pointing fractions are
\begin{equation}
\begin{aligned}
F_{c}^{\ell}
&=\frac{1}{N}\sum_{x\in\mathcal{X}}
\mathtt{Ind}(\alpha_{x}^{\ell,c}>0),
&&c\in\{\mathrm{par},\mathrm{sel}\},\\
F_{\mathrm{know}}^{\ell}
&=\frac{1}{N}\sum_{x\in\mathcal{X}}
\mathtt{Ind}(\gamma_{x}^{\ell}>0).
\end{aligned}
\label{eq:correct-pointing}
\end{equation}
$\mathtt{Ind}$ returns one when its condition holds and zero otherwise. The denominator is the number of questions. Routing pointing is a paired-request diagnostic; knowledge pointing tests proximity to the correct reference. These fractions measure information present, while $E^{\mathrm{ans}}$ and $E^{\mathrm{know}}$ measure the consequences of intervention.

\subsection{Layerwise Comparisons and Uncertainty}

In the original evaluation, each partition is scored separately. A \emph{stable} effect is positive in selection and both validation partitions. Hidden-state routing must also pass its marker-task calibration in those partitions. Curves average the two validation effects and shade the range between them. This range is not a confidence interval, and a non-stable value is not set to zero.

For the expanded evaluation, earlier and later layer sets are fixed separately for each model before testing the new countries. We average damages within each country pair and layer set before applying the random correction. We report 95\% percentile bootstrap intervals from 2,000 resamples of the 24 country pairs. All variants and conditions of a pair stay together. The same resamples estimate differences between components or layer sets. Appendix~\ref{app:expanded-evidence} specifies the aggregation and fixed sets. Fresh model-loading repeats check reproducibility; they do not increase the number of independent samples.

\section{Narrow-Routing Measurement Details}
\label{app:narrow-details}

This appendix makes the paired-task measurements reproducible by fixing their state coordinates, fitting sets, calibration rules, and uncertainty interpretation. The recorded-state location exception is distinguished from the intervention location, and the later subsections give the component formulas and controls.

\subsection{Model and Task Records}

The frozen runs use the released instruction-tuned checkpoints named in Section~\ref{sec:setup}: Qwen with 36 decoder layers, Llama with 28, and Gemma with 34. The archived contracts record the exact configuration, tokenizer, weight-index, and weight-shard hashes. Copies of these contracts accompany the figure sources. Computation uses Brain Floating Point 16 (bfloat16), a 16-bit floating-point representation, and all parameters remain fixed during the interventions.

One frozen continent-name prompt is:
\begin{quote}
\ttfamily\small
First country: Kenya\\
Second country: China\\
Which continent is the first country located in?\\
Answer with only the continent name.
\end{quote}
The alternative request replaces first with second. The adjective format asks for the adjectival form of the requested country's continent. In the code format, the input additionally supplies the mapping Africa = dax, Asia = wug, Europe = blicket, and asks for the requested country's continent code.

The separate calibration prompt supplies two records: Kenya has marker dax and China has marker fep. It asks which marker belongs to the first or second record. Calibration therefore measures selection of supplied information, while the main task requires the country--continent association from the model.

Each candidate string is appended to the rendered prompt and scored token by token using its own preceding tokens. Intervention is applied at the question-end position of the specified layer. Layerwise knowledge scores are read at that same position after subsequent layers.

\begin{table}[!htbp]
\centering
\small
\begin{tabular}{llllll}
\toprule
Model & Format & \multicolumn{3}{c}{Stable layers} & \multicolumn{1}{c}{Validation peaks} \\
 & & Parameter & Hidden-state & Knowledge & (P / H / K) \\
\midrule
Qwen & noun & 27--29, 31 & --- & 27--36 & 0.55 / 0.06 / 7.64 \\
Qwen & adjective & 30--31 & 28--30, 32 & 27--36 & 0.28 / 0.11 / 6.25 \\
Qwen & code & 27--33 & --- & 27--36 & 1.88 / 0.03 / 9.30 \\
Llama & noun & 15, 21, 24 & --- & 14--28 & 0.10 / 0.05 / 16.16 \\
Llama & adjective & 14--18, 20--21, 24 & --- & 13--28 & 0.18 / 0.03 / 9.75 \\
Llama & code & 11--12, 14--26 & --- & 14--28 & 0.41 / 0.00 / 8.49 \\
Gemma & noun & 17, 19--24 & --- & 18--34 & 1.96 / 0.29 / 19.96 \\
Gemma & adjective & 20--24 & --- & 18--34 & 3.83 / 0.15 / 22.49 \\
Gemma & code$^*$ & 24--26 & --- & 17--34 & 0.41 / $-$0.01 / 10.86 \\
\bottomrule
\end{tabular}
\caption{Stable 1-based layers and peak answer-margin effects. All three peak entries are answer effects: parameter routing / transferred object selection / target content. They may be compared within a model--format row, but raw magnitudes are not pooled or ranked across models. ``---'' means that no layer has a positive effect in the screening split and both validation splits. The peak column still reports the largest validation mean, including non-stable layers; selection stability additionally requires marker-task calibration. $^*$Gemma code answers fail the behavioral gate in one split and are diagnostic only.}
\label{tab:lifecycle}
\end{table}

\subsection{Fitting and Centering}
\label{app:fitting-formulas}

For each model, layer, and answer format, let $\mathcal{X}_{\mathrm{fit}}$ be its original fitting questions. The fit mean is $\vect{\mu}^{\ell}=|\mathcal{X}_{\mathrm{fit}}|^{-1}\sum_{x\in\mathcal{X}_{\mathrm{fit}}}\vect{h}_{x}^{\ell}$. For each continent $k\in\mathcal{K}$, let $\mathcal{X}_{\mathrm{fit},k}$ contain the fitting questions whose requested country's continent is $k$. Then $\vect{\mu}_{k}^{\ell}=|\mathcal{X}_{\mathrm{fit},k}|^{-1}\sum_{x\in\mathcal{X}_{\mathrm{fit},k}}\vect{h}_{x}^{\ell}$. There are six questions in the full fit and two in each continent group, all from the current answer format. We stack the centered continent means into the matrix
\begin{equation}
\vect{V}^{\ell}
=\bigl[\vect{\mu}_{\mathrm{Africa}}^{\ell}-\vect{\mu}^{\ell},\
       \vect{\mu}_{\mathrm{Asia}}^{\ell}-\vect{\mu}^{\ell},\
       \vect{\mu}_{\mathrm{Europe}}^{\ell}-\vect{\mu}^{\ell}\bigr]^\top
\in\mathbb{R}^{3\times H}.
\label{eq:knowledge-fit}
\end{equation}
Its first two right singular vectors form the columns of $\vect{U}^{\ell}$. Thus $\vect{P}^{\ell}=\vect{U}^{\ell}(\vect{U}^{\ell})^{\top}$ and the continent references are $\vect{\kappa}_{k}^{\ell}=\vect{P}^{\ell}(\vect{\mu}_{k}^{\ell}-\vect{\mu}^{\ell})$. These are separate fits for each answer format, not a shared coordinate system across formats.

For request-conditioned means, superscripts $\mathrm{main}$ and $\mathrm{cal}$ indicate the main continent task and auxiliary calibration task, respectively; subscripts $\mathrm{first}$ and $\mathrm{second}$ identify the requested position. All means below use the corresponding task's fit partition. The raw directions are
\begin{equation}
\begin{aligned}
\vect{r}^{\ell,\mathrm{raw}}
&=\vect{\mu}_{\mathrm{first}}^{\ell,\mathrm{main}}
-\vect{\mu}_{\mathrm{second}}^{\ell,\mathrm{main}},\\
\vect{o}^{\ell,\mathrm{raw}}
&=\vect{\mu}_{\mathrm{first}}^{\ell,\mathrm{cal}}
-\vect{\mu}_{\mathrm{second}}^{\ell,\mathrm{cal}}.
\end{aligned}
\label{eq:raw-routes}
\end{equation}
First remove the knowledge projection:
$\widetilde{\vect{r}}^{\ell}=(\vect{I}_{H}-\vect{P}^{\ell})\vect{r}^{\ell,\mathrm{raw}}$ and
$\widetilde{\vect{o}}^{\ell}=(\vect{I}_{H}-\vect{P}^{\ell})\vect{o}^{\ell,\mathrm{raw}}$.
The tildes mark these knowledge-excluded candidates. For a vector $\vect{v}\in\mathbb{R}^{H}$, define $\mathtt{Unit}(\vect{v})=\vect{v}/\|\vect{v}\|_2$ when its norm exceeds $10^{-12}$, and the zero vector otherwise. Remove overlap with the other candidate and normalize:
\begin{equation}
\begin{aligned}
\vect{r}^{\ell,\mathrm{global}}
&=\mathtt{Unit}\!\left(
\widetilde{\vect{r}}^{\ell}
-\bigl[\mathtt{Unit}(\widetilde{\vect{o}}^{\ell})^\top
\widetilde{\vect{r}}^{\ell}\bigr]\mathtt{Unit}(\widetilde{\vect{o}}^{\ell})
\right),\\
\vect{o}^{\ell}
&=\mathtt{Unit}\!\left(
\widetilde{\vect{o}}^{\ell}
-\bigl[\mathtt{Unit}(\widetilde{\vect{r}}^{\ell})^\top
\widetilde{\vect{o}}^{\ell}\bigr]\mathtt{Unit}(\widetilde{\vect{r}}^{\ell})
\right).
\end{aligned}
\label{eq:unique-routes}
\end{equation}
Each residual is orthogonal to the other candidate \emph{before that other candidate is residualized}. The two final unit directions therefore need not be mutually orthogonal, and deleting one does not leave the fitted coefficient of the other unchanged. We use them as two separately defined measurement directions, not as an additive orthogonal decomposition of the hidden state.

Routing deletion uses the fit-derived center
$\vect{b}^{\ell}=\tfrac12(\vect{\mu}_{\mathrm{first}}^{\ell,\mathrm{main}}+\vect{\mu}_{\mathrm{second}}^{\ell,\mathrm{main}})$;
knowledge deletion uses $\vect{\mu}^{\ell}$, as specified in Equation~\ref{eq:deletions}. The balanced requests make these centers equal in exact arithmetic. They are never chosen separately for validation pairs.

For natural routing pointing, let $\bar{x}$ be the question with the same country order and answer format but the opposite request. Its center is
$\vect{b}_{x}^{\ell,\mathrm{pair}}=\tfrac12(\vect{h}_{x}^{\ell}+\vect{h}_{\bar{x}}^{\ell})$.
This pair-specific center is used only for the sign diagnostic in Equation~\ref{eq:correct-pointing}, not for deletion. The diagnostic compares two unmodified requests; it is not a standalone classifier given only one question.

\paragraph{Final-layer extraction.}
The original profiles fit and read states returned by the model's hidden-state interface. Its final entry is after the final normalization, while intervention is applied to the final decoder-block output before normalization. The expanded evaluation aligns fitting, readout, and intervention at decoder-block outputs before final normalization. Nonfinal extraction locations agree. Consequently, the original and expanded final-layer points use different state coordinates. The expanded full-depth results provide the location-aligned final-layer comparison.

\subsection{Auxiliary Calibration and Controls}

The auxiliary input explicitly assigns unrelated markers to two records. Shifting the question-end state toward the opposite request should change the preferred marker. A layer passes calibration in a partition if baseline marker accuracy is at least 90\%, directional switching succeeds on at least two-thirds of questions, and the best of eight matched random controls switches at most one-third. Stable main-task hidden-state-routing effects additionally require calibration in selection and both validation partitions.

For each main-task deletion, eight standard-normal vectors are independently generated, normalized, and scaled to that deletion's Euclidean length. They are unrestricted random changes, not projections removed from a learned random subspace. Seeds combine a fixed experiment seed with the record identifier. Each control is evaluated on the same question, layer, and position as its corresponding deletion.

Main curves use Equation~\ref{eq:answer-effect}: they average each control across questions and then take the maximum. The archived, more conservative alternative instead takes the maximum within each question:
\begin{equation}
E_{c}^{\ell,\mathrm{per\mbox{-}question}}
=\frac1N\sum_{x\in\mathcal{X}}
\left(\Delta m_{x}^{\ell,c}
-\max_{j=1,\ldots,8}\Delta m_{x}^{\ell,c,j}\right).
\label{eq:per-question-control}
\end{equation}
The superscript identifies the per-question maximum. This value cannot exceed $E_{c}^{\ell,\mathrm{ans}}$, and the two are not interchangeable. The analogous distinction applies to knowledge-score losses.

The code format supplies an output mapping, but not the queried country--continent association. Correctness is based on the two candidate strings' length-averaged log probabilities, not on unrestricted generation accuracy.

\subsection{Replication and Uncertainty}

Each evaluated partition has six questions from three country pairs. The two requests for one pair are related observations, not six independent country pairs. The fit, selection, and two validation partitions have disjoint countries. Each run is repeated after reloading the model in a fresh process. The archived comparisons report 5,292 paired component records and 15,876 paired downstream-knowledge records, with no scientific-field mismatches. Input rendering, token positions, and file hashes are retained for reproducibility.

Figures use only the primary runs; reloaded copies are replication checks, not additional samples. Lines average the two validation-partition values at each layer, and shaded ranges show their minimum and maximum. No bootstrap confidence intervals or statistical significance claims are attached to these ranges. Stable layers satisfy the cross-partition positivity criterion of Section~\ref{sec:setup}. A non-stable layer is not evidence of exactly zero effect.

The unique routing effects exclude directions shared with the other fitted components. The three deletion effects need not add to the effect of deleting their union because subsequent computation is nonlinear. They are reported as changes in answer margin, not fractions of a total control budget.

\begin{figure}[!ht]
\centering
\includegraphics[width=\linewidth]{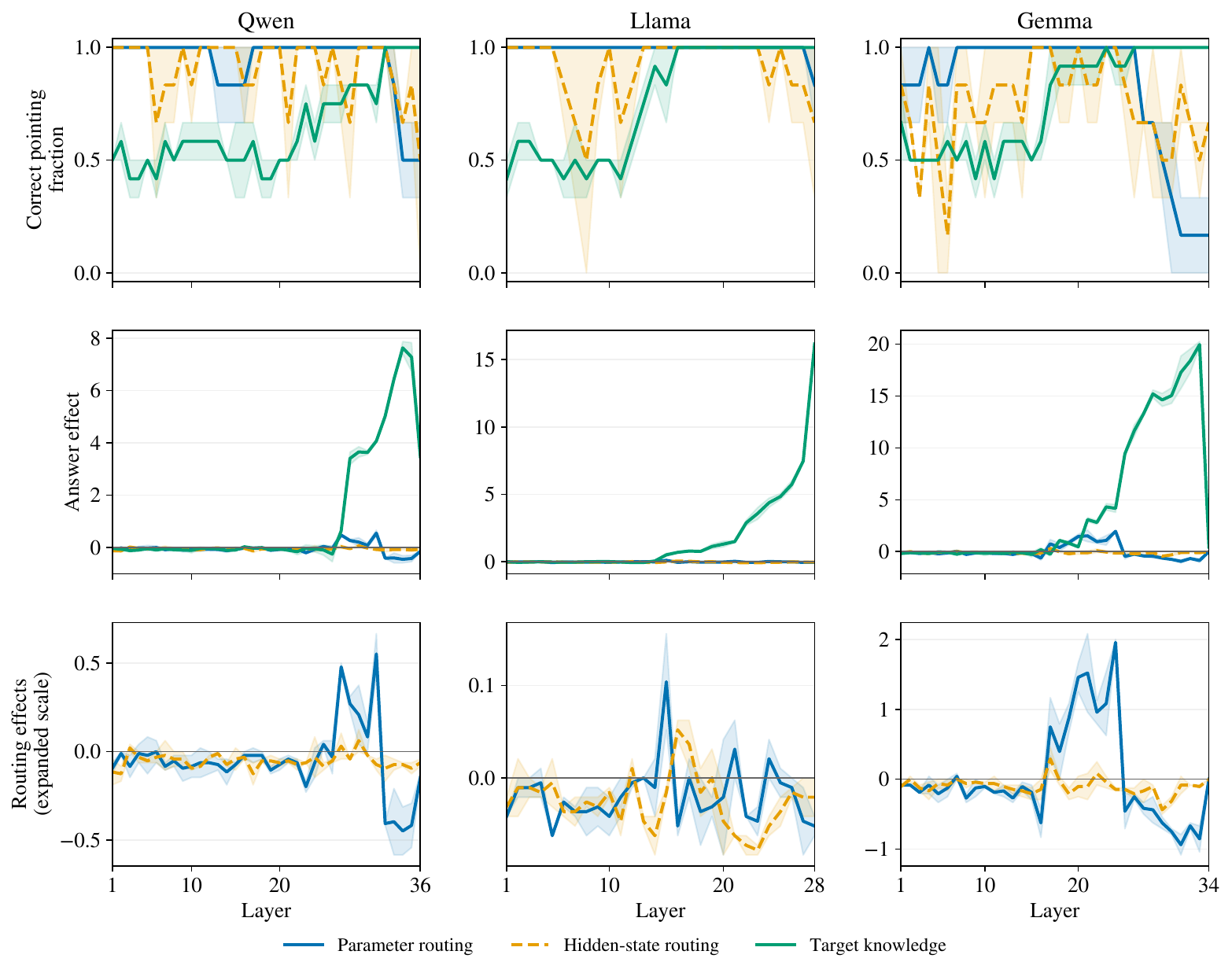}
\caption{\textbf{Three-component profiles on continent-name questions.} Top: correct-pointing fractions. Middle: random-corrected deletion effects on the answer. Bottom: the same routing effects with an expanded vertical scale, without normalization. Lines average validation A and B; shading spans their two values. Orange dashed selection curves include unqualified layers. The presence of a curve does not assert a stable effect.}
\label{fig:lifecycle}
\end{figure}

\begin{figure}[p]
\centering
\includegraphics[width=\linewidth]{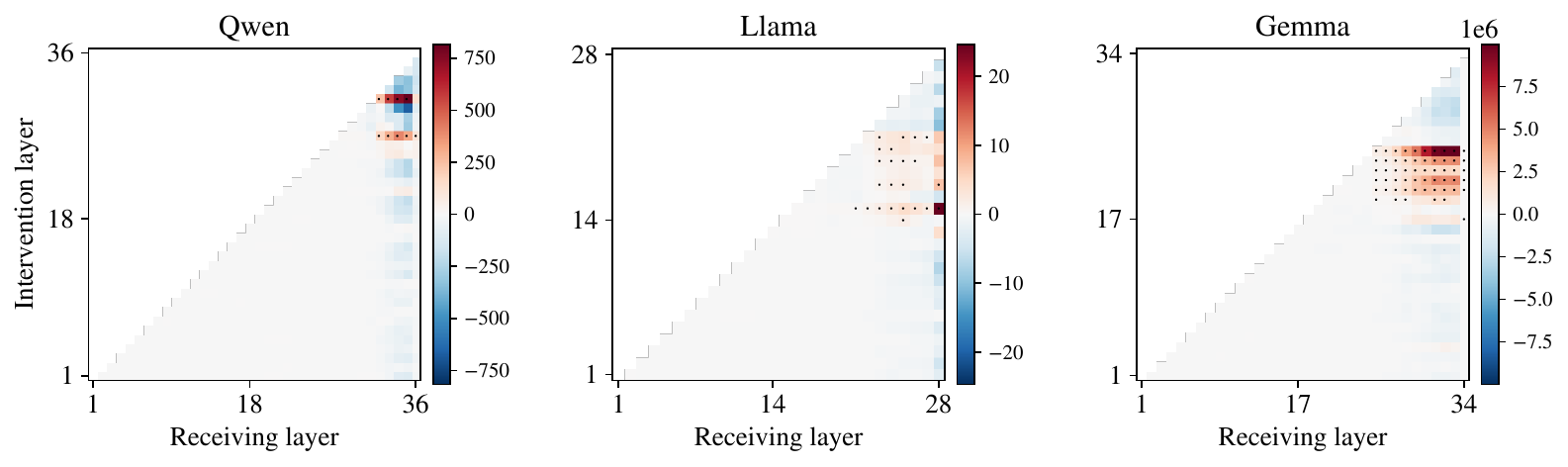}
\caption{\textbf{Paired-task parameter-routing deletion affects later knowledge.} Columns show the continent-name task in each model. Color is the mean random-corrected knowledge loss across the two validation partitions. Black dots mark layer pairs positive in selection and both validation partitions, not statistical significance. Blank cells have no later receiving layer. Model color scales differ because raw knowledge scores are not directly comparable.}
\label{fig:later-knowledge}
\end{figure}

\begin{figure}[p]
\centering
\includegraphics[width=\linewidth]{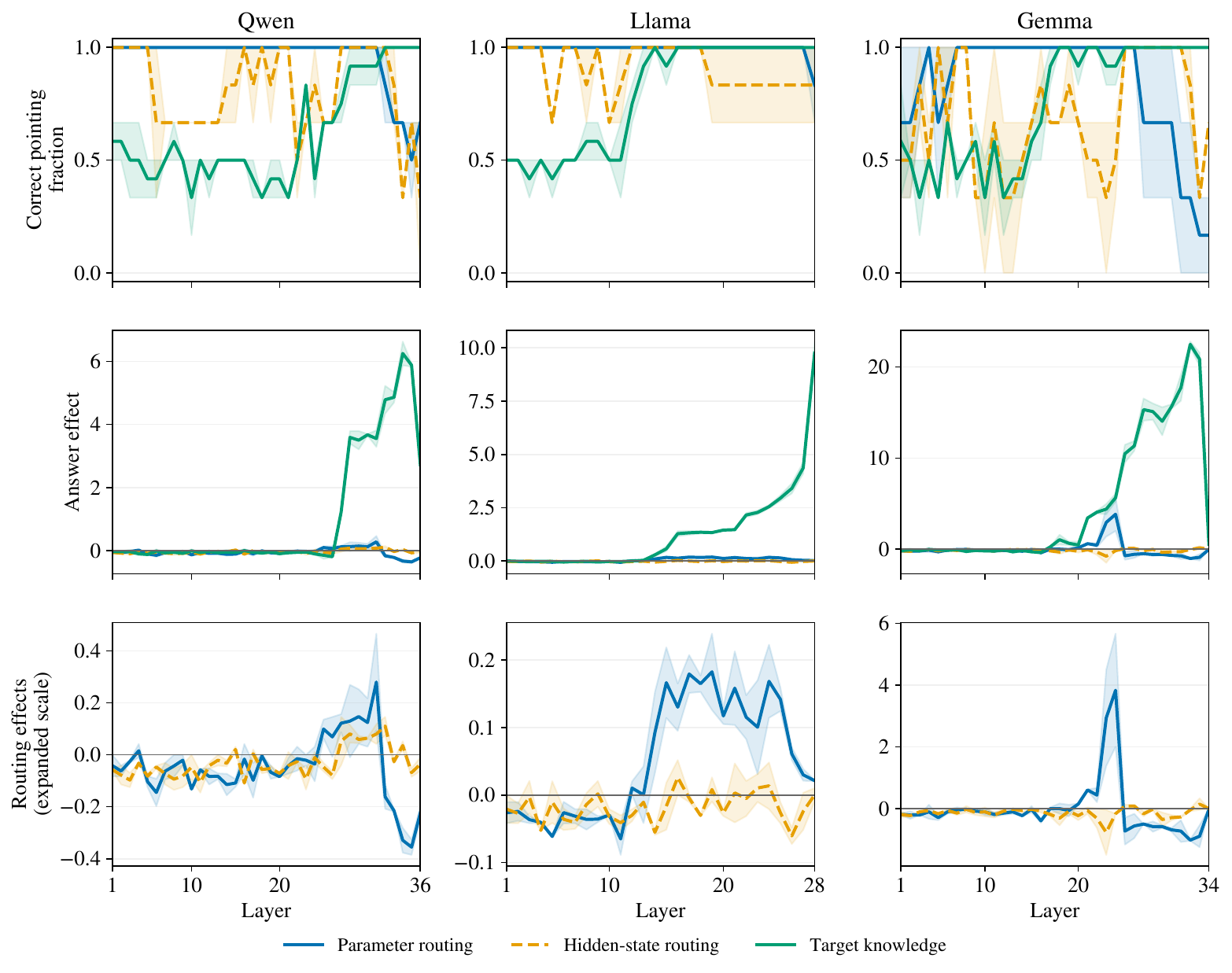}
\caption{\textbf{Adjective-format lifecycle.} Layout and measurements follow Figure~\ref{fig:lifecycle}. Lines average the two validation results; shading is their range, not a confidence interval. The bottom row expands the routing scale without normalization. Dashed orange lines show the transferred selection candidate, including unqualified layers. Stable effects require the additional criteria in Section~\ref{sec:setup}; readable or positive values alone do not qualify a layer.}
\label{fig:lifecycle-adjective}
\end{figure}

\begin{figure}[p]
\centering
\includegraphics[width=\linewidth]{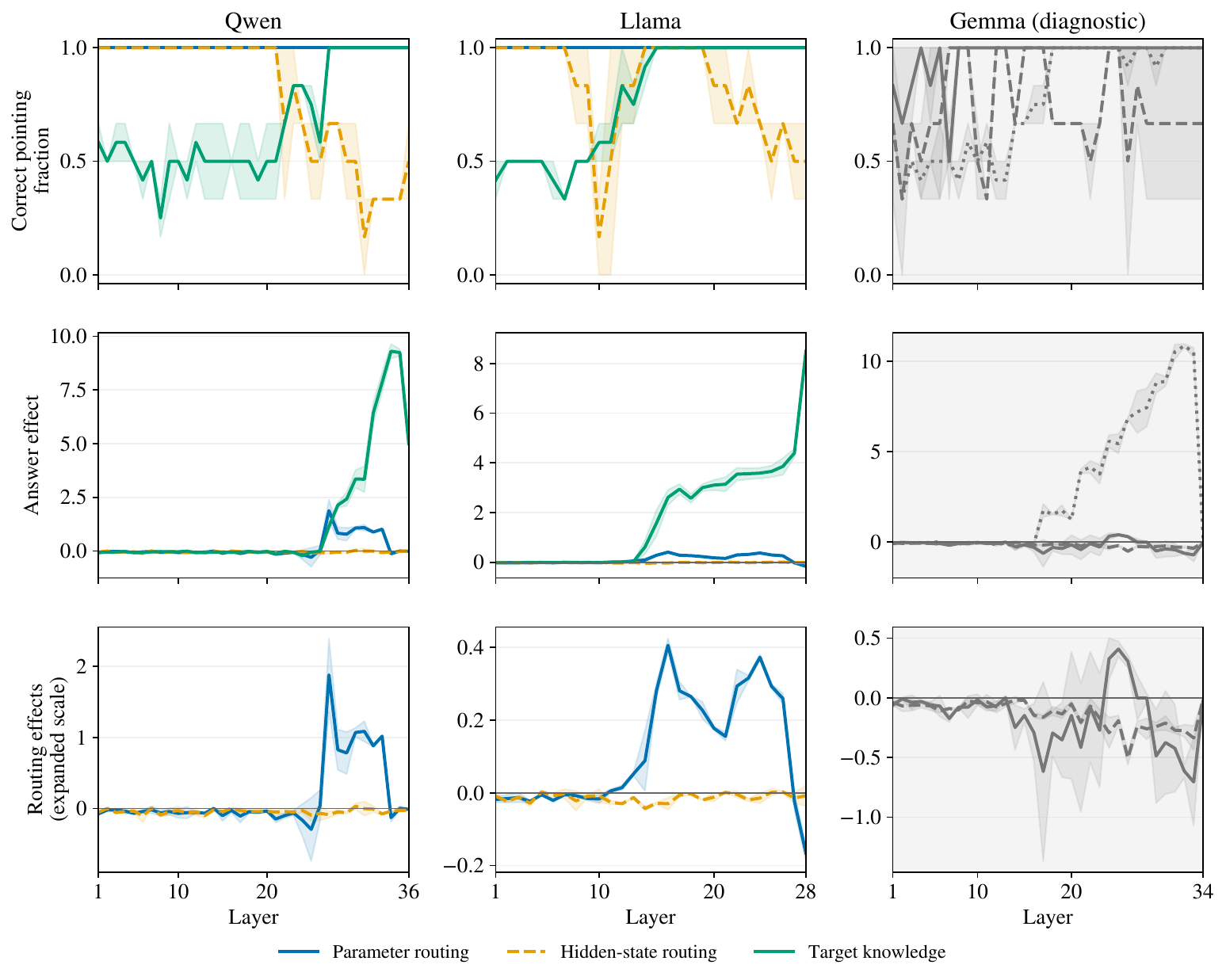}
\caption{\textbf{Code-format lifecycle.} Layout and measurements follow Figure~\ref{fig:lifecycle}. Gemma's gray column is diagnostic because this model--format combination failed behavioral qualification. Within that column, solid, dashed, and dotted lines denote parameter routing, hidden-state routing, and target knowledge, respectively. Shading is the range of the two validation results, not a confidence interval. Diagnostic values are retained rather than treated as confirming the main conclusions.}
\label{fig:lifecycle-code}
\end{figure}

Figures~\ref{fig:lifecycle}, \ref{fig:lifecycle-adjective}, and \ref{fig:lifecycle-code} give all three components together for each answer format. Figure~\ref{fig:later-knowledge} reports the original parameter-routing effects on later knowledge.

\FloatBarrier
\section{Expanded Evaluation and Repeated Interventions}
\label{app:expanded-evidence}

\subsection{New Countries and Fixed Layer Comparisons}

The expanded evaluation tests transfer to 48 countries excluded from the original fitting and evaluation partitions. They form 24 non-overlapping country pairs, each with countries from different continents. Swapping country order, rephrasing the question, and requesting either country gives eight variants per pair. All 192 questions prefer the correct candidate in every model. We construct the continent-name knowledge space and both routing directions from the original fitting questions using the same formulas; no expanded question is used for fitting.

The earlier and later layer sets were fixed from the original evidence before expanded evaluation. Qwen uses layers 29 and 31 versus 32 and 34; Gemma uses layer 24 versus layers 25 and 30; Llama uses layers 15 and 24 versus layer 28. These are comparison sets, including nonadjacent layers, rather than assertions of continuous processing stages. Expanded fitting, readout, and interventions all use decoder-block outputs before the final normalization. The final-layer measurement therefore refers to the block output, not the normalized state returned by some model interfaces.

The independent resampling unit is a country pair, not one of its eight question variants. Let $\mathcal{Q}$ contain the 24 country pairs and $\mathcal{X}_{q}$ the eight questions for pair $q\in\mathcal{Q}$. For a selected layer set $\mathcal{L}$, define the pair-averaged true and random damages by
\begin{equation}
\begin{aligned}
\overline{\Delta m}_{q,\mathcal{L}}^{c}
&=\frac{1}{8|\mathcal{L}|}
\sum_{x\in\mathcal{X}_{q}}\sum_{\ell\in\mathcal{L}}
\Delta m_{x}^{\ell,c},\\
\overline{\Delta m}_{q,\mathcal{L}}^{c,j}
&=\frac{1}{8|\mathcal{L}|}
\sum_{x\in\mathcal{X}_{q}}\sum_{\ell\in\mathcal{L}}
\Delta m_{x}^{\ell,c,j}.
\end{aligned}
\label{eq:pair-aggregation}
\end{equation}
The bars denote averages over that pair's variants and selected layers. The pooled answer effect used in Equation~\ref{eq:handoff-contrasts} is
\begin{equation}
E_{\mathcal{L},c}^{\mathrm{ans}}
=\frac1{24}\sum_{q\in\mathcal{Q}}\overline{\Delta m}_{q,\mathcal{L}}^{c}
-\max_{j=1,\ldots,8}\frac1{24}
\sum_{q\in\mathcal{Q}}\overline{\Delta m}_{q,\mathcal{L}}^{c,j}.
\label{eq:pooled-effect}
\end{equation}
For each of 2,000 bootstrap resamples, we sample 24 pairs with replacement and repeat this entire aggregation, including the maximum over controls. All eight variants and all compared conditions of a sampled pair stay together. Using the same resampled pairs for both terms gives the interval for each contrast in Equation~\ref{eq:handoff-contrasts}. These are percentile intervals for the specified comparisons, not simultaneous confidence bands over all layers.

\begin{figure}[p]
\centering
\includegraphics[width=\linewidth]{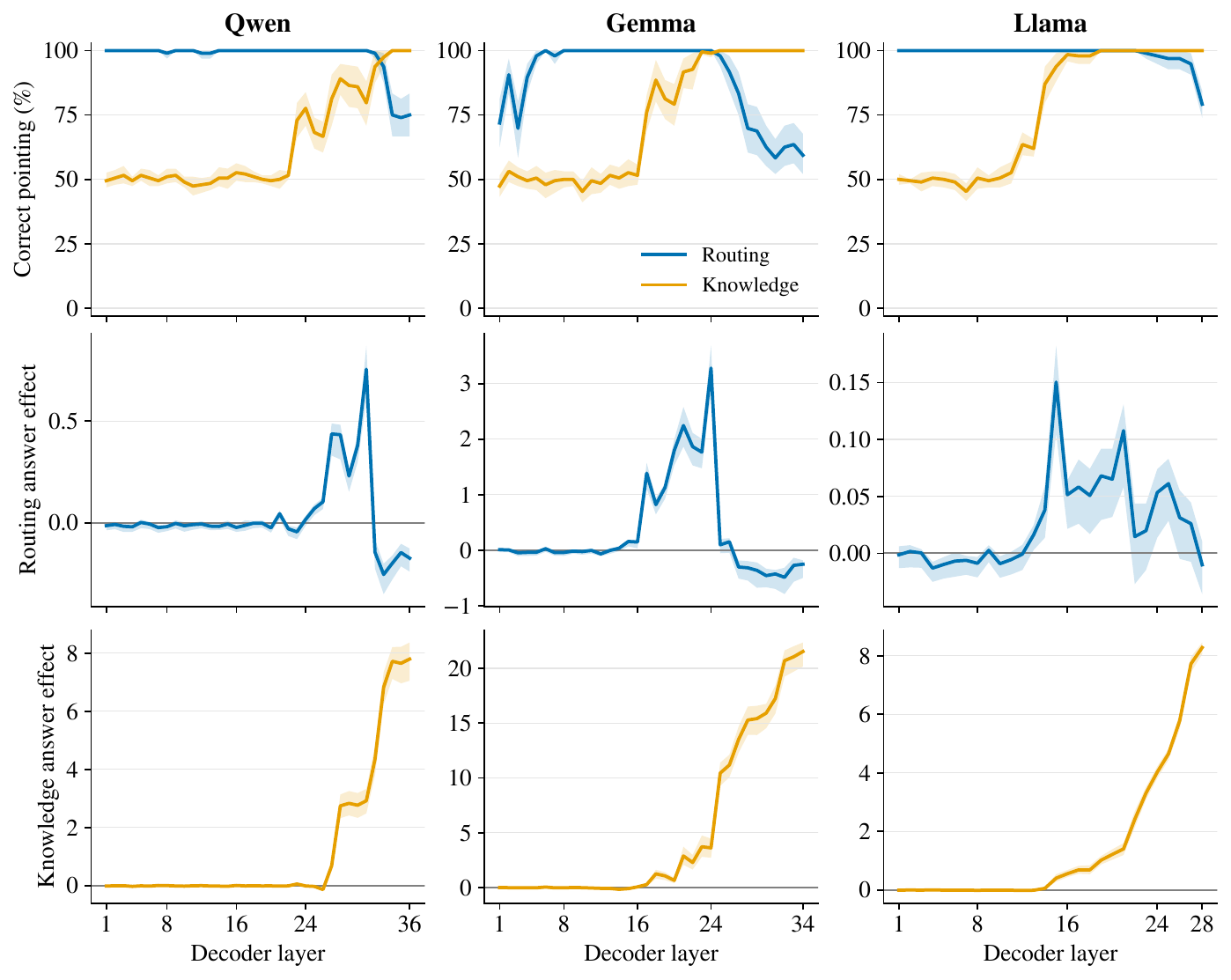}
\caption{\textbf{Expanded depth profiles on 24 new country pairs.} Top: paired routing correct-pointing fractions and target-knowledge correct-pointing fractions. Middle: routing answer effects. Bottom: target-knowledge answer effects. Shading gives pointwise 95\% country-pair bootstrap intervals; it differs from the two-partition ranges in Figure~\ref{fig:lifecycle}. Each column uses its model's full decoder depth and separately fitted components. Routing and knowledge effects have separate vertical scales.}
\label{fig:expanded-full-depth}
\end{figure}

Table~\ref{tab:expanded-windows} gives the exact estimates. Figure~\ref{fig:expanded-full-depth} shows the full expanded profiles. Later-layer pointing averages the paired sign diagnostic over the fixed later layer set; it measures request separation and is not an independent classifier for an isolated question.

\begin{table}[!htbp]
\centering
\footnotesize
\setlength{\tabcolsep}{3pt}
\begin{tabular}{lccc}
\toprule
Comparison & Qwen & Gemma & Llama \\
\midrule
Earlier routing & $0.515\;[0.403,0.577]$ & $3.280\;[2.853,3.681]$ & $0.102\;[0.075,0.120]$ \\
Later routing & $-0.155\;[-0.215,-0.113]$ & $-0.148\;[-0.299,-0.057]$ & $-0.010\;[-0.036,0.010]$ \\
Routing decline & $0.669\;[0.560,0.758]$ & $3.428\;[3.031,3.846]$ & $0.112\;[0.075,0.144]$ \\
Later knowledge & $6.035\;[5.583,6.438]$ & $13.270\;[12.365,13.889]$ & $8.281\;[8.053,8.453]$ \\
Knowledge advantage & $6.190\;[5.738,6.599]$ & $13.418\;[12.518,14.065]$ & $8.291\;[8.066,8.467]$ \\
\bottomrule
\end{tabular}

\caption{Expanded answer effects and 95\% country-pair bootstrap intervals. Routing decline is the earlier-minus-later routing effect; knowledge advantage is the later knowledge effect minus the later routing effect. Each model uses the same 24 new country pairs and its fixed layer sets. Negative corrected effects mean less damage than the strongest tested random control, not exactly zero raw deletion damage.}
\label{tab:expanded-windows}
\end{table}

Each complete experiment is repeated after fresh model loading. Primary and repeated answer margins agree exactly for all 6,912 Qwen, 6,528 Gemma, and 5,376 Llama question--layer records, including every intervention condition. Repeated runs are reproducibility checks and are not pooled as additional observations.

\subsection{Balanced Wrong-Label Directions}

This control tests whether Qwen's earlier routing effect depends on the true request labels. The six original fitting examples admit ten distinct balanced three-versus-three divisions when label reversal is treated as equivalent. One is the true first-versus-second division; the remaining nine define wrong-label controls. For every division we recompute the request-mean difference, remove its fitted knowledge projection and its overlap with the knowledge-excluded auxiliary selection direction, and normalize the result.

Natural-size deletion removes each wrong direction's coefficient from the centered state. Length-matched deletion rescales that vector to the true routing deletion's Euclidean length before insertion into the model. This is vector-length matching; the post-rounding length search described below is used only for repeated random paths. We compare the true direction with the largest mean damage across all nine wrong-label directions. The maximum is recomputed within each country-pair bootstrap resample.

Both comparisons pass in the fixed earlier set, layers 29 and 31 (Figure~\ref{fig:expanded-controls}a). Examined separately, layer 29 has an interval spanning zero for both comparisons; layer 31 has positive intervals. We retain these single-layer results in the figure data. Two complete runs contain 768 question--layer records each, with identical margins for all wrong-label conditions.

\subsection{Repeated Deletion and Random Paths}

At the question-end position, we remove the routing coefficient from each selected layer's output along that layer's fitted direction. The coefficient is recomputed from the state produced by all previous interventions. Target-knowledge deletion analogously removes the current knowledge-space projection. We test single deletions at layers 29 or 32 and repeated deletions at every layer from 29--36 or 32--36. No intervention is applied at later answer-token positions.

Eight Gaussian random paths control for accumulated perturbation. At each active layer, a path's perturbation is matched to the actual change on the corresponding true-deletion trajectory. For the repeated routing comparison, we adjust its scale to match the length after bfloat16 rounding, without using answer or knowledge scores to select the scale. The search uses 129 logarithmically spaced scales from 0.01 to 100 times the target length, followed by two local refinements with 33 scales each. For repeated knowledge deletion, the original controls already meet the length criterion.

We require relative length error at most 5\% at at least 95\% of the active positions. After rounding-aware matching, only 12 of 12,288 earlier-routing positions and 5 of 7,680 later-routing positions exceed the error threshold. The original routing controls exceeded it at about 8--9\% of positions; Figures~\ref{fig:expanded-controls}b,c use the corrected controls. We also check the component remaining after deletion. Its length is divided by the largest of the incoming length, one tenth of its mean fitting-example length, and a numerical floor of $10^{-12}$. We require this ratio to be at most 5\% at at least 95\% of active positions. The target deletions satisfy this criterion.

To test whether repeated later routing deletion adds a practically meaningful loss, let $\overline{\Delta m}_{q}^{\mathrm{repeat}}$ and $\overline{\Delta m}_{q}^{\mathrm{single}}$ be the raw damages averaged over pair $q$'s eight variants for deletion at layers 32--36 and at layer 32 alone, respectively. Let $\overline{m}_{q}^{0}$ be that pair's mean unmodified answer margin. The relative added loss is
\begin{equation}
\rho=\frac1{24}\sum_{q\in\mathcal{Q}}
\frac{\overline{\Delta m}_{q}^{\mathrm{repeat}}
-\overline{\Delta m}_{q}^{\mathrm{single}}}
{\overline{m}_{q}^{0}}.
\label{eq:relative-added-loss}
\end{equation}
The denominator is positive because all baseline questions pass qualification. This quantity uses raw deletion damage, not the random-corrected effect. Its one-sided 95\% bootstrap upper bound is compared with the 10\% bound fixed before this experiment. Repeated target-knowledge and routing effects use the same pair resamples when their difference is estimated.

\subsection{Additional Knowledge-Relation Screening}

A separate extension considered works' original languages, with English, Tamil, and French labels checked against public records. It stopped before component fitting because the frozen candidate pool did not yield disjoint pairs spanning all three language combinations with every question variant correct. Across 768 screening questions, candidate accuracy was 52.3\%, with large differences among language labels. Consequently, this extension supplies no intervention evidence about the handoff. The expanded mechanism results above remain about country--continent knowledge.

\section{Cross-Format and Fitting-Robustness Checks}
\label{app:format-robustness}

\subsection{Fixed Knowledge Changes across Answer Forms}

We construct the target and alternative continent references from the noun-answer fit partition. Their difference, Equation~\ref{eq:content-transfer}, is applied to the same-layer question-end state of noun, adjective, or code questions. We evaluate Qwen at layer 34, Llama at layer 28, and Gemma at layer 33, fixed before this comparison. The noun direction is never fitted again on an adjective or code response.

Each model tests all six code mappings. Baseline qualification is checked separately for each mapping; mappings with a wrong baseline answer remain diagnostic. A transfer passes if its random-corrected shift is positive in both validation partitions and at least two of the three country pairs in each partition have a positive pair mean. The reported shift first averages the six questions in each partition and then averages the two partition results. It is not a percentage-point change in accuracy. Both natural answer forms pass in all three models. Code transfer passes for 1/5 qualified mappings in Qwen, 0/4 in Llama, and 0/2 in Gemma.

\subsection{Fit and Random-Seed Sensitivity}

Four fits use either all three original fitting pairs or omit one pair at a time. Each is tested with three random seeds on both validation partitions. A fit passes when its effect is positive in every seed--partition combination. The registered aggregate criterion requires at least three of the four fits.

At the earlier comparison layers, Qwen passes with 3/4 fits at layers 29 and 31, and Gemma passes with 4/4 at layer 24. Their later routing comparisons pass with 0/4 fits, while content passes with 4/4 fits at every listed comparison layer. Llama routing passes with only 2/4 fits at layers 15 and 24, below the same criterion; its content effect passes with 4/4. These checks concern sensitivity to fitting examples and random seeds. The 24-pair evaluation separately addresses sensitivity to the evaluation countries.

\subsection{Raw Damage, Random Damage, and Deletion Length}

For Qwen, the transition from layer 31 to 32 changes mean route-deletion damage from 0.8021 to 0.0104 and mean strongest-random damage from 0.2500 to 0.4167. Meanwhile, deletion length increases from 11.2052 to 12.0985. For Gemma, the layer-24 to layer-25 transition changes true damage from 3.2083 to 0.2292 and random damage from 1.2500 to 0.6875. These decompositions support a decrease in true deletion damage, rather than attributing the corrected-score decline only to a changing random baseline.

The three-model supplement includes 20,832 intervention records across primary and fresh-load runs. All 10,416 paired answer-margin records agree. The synchronized archive contains 93 files with matching hashes. Fresh-load repeats check computational reproducibility and are not counted as additional independent observations.

\section{Route Strength and Cross-Model Timing Diagnostics}
\label{app:route-strength-timing}

\subsection{Why the Paired Three-Model Records Are Not a Direct Replication}

Section~\ref{sec:parameter-routing} measures Qwen on natural single-country questions with one route direction fitted separately for each country pair. Its route-strength denominator is the full hidden-state length. The three-model experiment instead asks paired-country questions, uses one global first-request-minus-second-request direction shared by all evaluated pairs, and divides the natural signed coefficient by half of the fit requests' separation along that direction. A normalized paired-task coordinate of one therefore means that the validation state is displaced by roughly one fitted-request half-separation. It is not the fraction of a hidden state occupied by routing and cannot be compared numerically with $S^{\mathrm{route}}$ in Section~\ref{sec:parameter-routing}.

\begin{figure}[t]
\centering
\includegraphics[width=\linewidth]{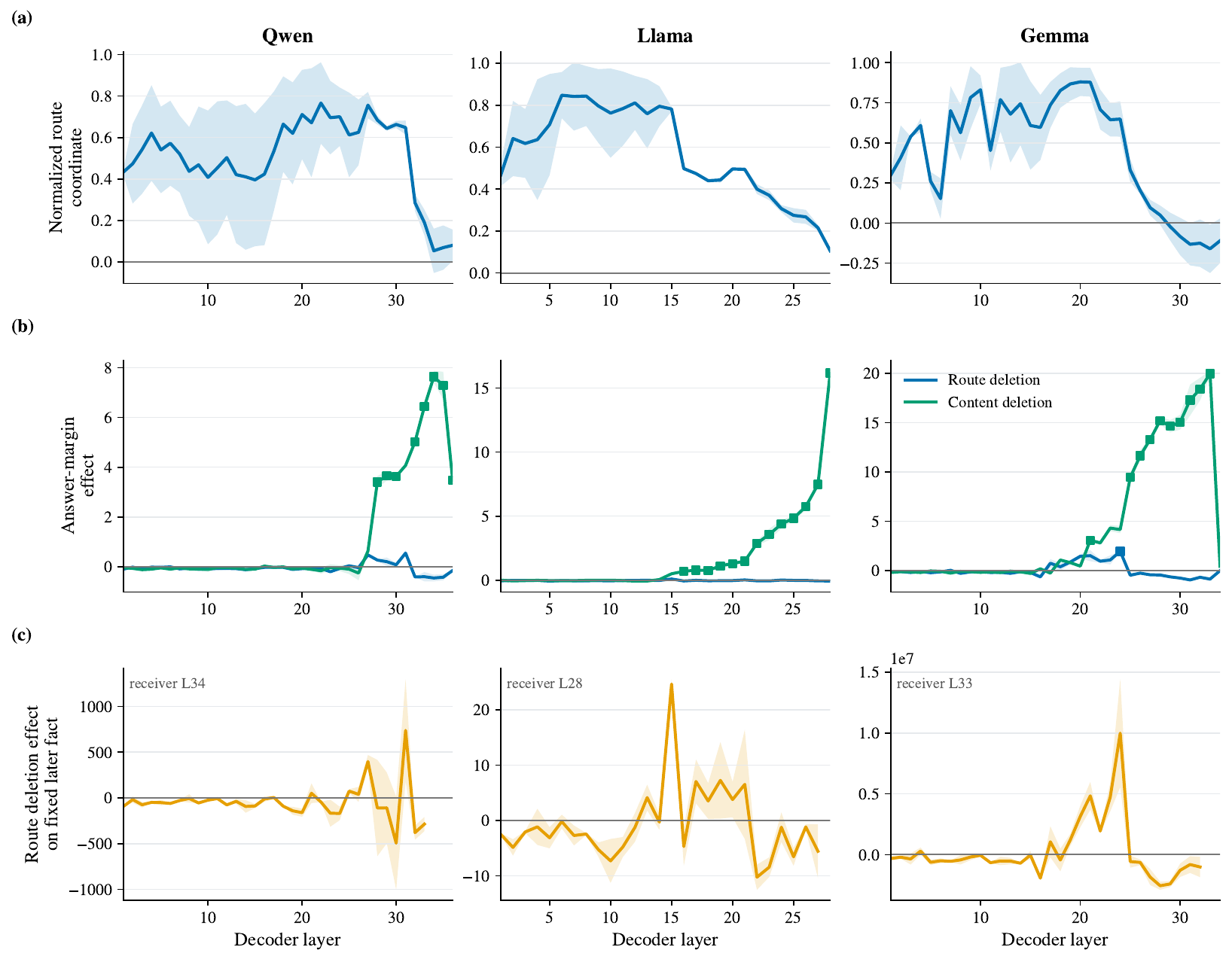}
\caption{\textbf{The paired-country records contain related ingredients but not one cross-model route-first schedule.} Columns show Qwen, Llama, and Gemma for continent-name answers. (a) The unmodified state's signed coordinate along the global request direction, normalized by the fitted request half-separation; values are interpreted within a model and are not whole-state percentages. (b) Answer-margin effects of deleting the global request coordinate or fitted continent content, each relative to eight random changes matched to its actual deletion length. Squares mark layers passing the same split-level stability rule in both validation partitions. (c) The effect of deleting the global route at each source layer on a fixed later fitted-fact score; the receivers are layers 34, 28, and 33 for Qwen, Llama, and Gemma. No continent-name source layer passes the fixed-later-fact stability rule in both partitions. The figure is a diagnostic reanalysis of an existing paired-question protocol, not a direct replication of the Qwen natural-question timing experiment.}
\label{fig:three-model-paired-trajectories}
\end{figure}

The three-model analysis retains all three answer forms and both validation partitions. Each model--format combination has six validation country pairs and 12 questions. Lines in Figure~\ref{fig:three-model-paired-trajectories} average validation A and B; shading spans their two values. A square marks a layer only when both partitions have positive random-corrected effect and at least 80\% of their individual questions have positive targeted-minus-random damage. Primary and fresh-load runs agree exactly for 5,292 component records and 15,876 source-to-receiver knowledge records; the repeat is a computational check, not an additional sample.

\subsection{Model-Specific Trajectories}

\paragraph{Qwen.}
The global paired-request coordinate enters a higher mid-layer range before fitted-content deletion becomes stably consequential for continent-name answers at layers 28--30 and 32--36. The global-route deletion itself does not pass the strict split-level answer gate at any continent-name layer, nor does it pass the fixed-later-fact gate. Thus the paired protocol complements, but does not reproduce, the natural-question result: the clean route-strength-to-later-fact sequence remains specific to the pair-conditioned Qwen analysis in Section~\ref{sec:parameter-routing}.

\paragraph{Llama.}
The normalized global-request coordinate is already high in early layers, remains high through roughly layer 15, and then declines. Route deletion has no sustained stable answer window for noun or adjective answers; the code task has only two isolated passing layers, 15 and 24. Fitted-content deletion becomes stably consequential from layer 16 for noun answers and layer 15 for adjective and code answers. No source passes the fixed layer-28 fact gate. Llama therefore does not show the same clean route-strength-before-route-to-fact trajectory.

\paragraph{Gemma.}
The global-request coordinate has a broad middle-layer high region. For continent-name answers, fitted-content deletion passes at layer 21 and then continuously from layers 25--33, while global-route deletion passes only at layer 24. This is a partial overlapping analogue: the request coordinate is strong before the local route-answer effect, but content is already becoming consequential and the route does not acquire a sustained fixed-later-fact effect. The adjective and code formats retain sustained content effects without a sustained route-effect window.

Across the three models, the shared result is therefore a distinction among natural coordinate strength, answer dependence, and later-fact dependence, not a universal ordering. Qwen supplies the clearest natural-question route-leading trajectory; Gemma supplies a more overlapping partial analogue in the paired protocol; Llama is a boundary case with dispersed routing effects and earlier sustained content dependence. The plotted values and source hashes are included under \path{anc/evidence/route_strength_timing/}.

\FloatBarrier
\section{Common-Protocol Direction and Direct-Selection Checks}
\label{app:same-condition}

The common-protocol comparison asks which part of the late effect changes when the input task is held fixed and the request measurement changes. We first define the direction and center choices and their aggregation, then give the direct-selection readouts and stopping rules used to test a different function: choosing between usable candidate facts.

\subsection{Direction Construction and Centering}

This comparison tests whether a late-layer contrast requires a task change. We hold Qwen-2.5-3B-Instruct weights, float32 computation, final-input decoder-block outputs, answer scoring, and the 24 validation country pairs fixed. The paired templates are ``Which continent is the first country located in?'', ``On which continent is the first country located?'', and ``What continent is the first country in?'', following the same two-country listing and continent-name instruction. Replacing first with second gives the opposite request. Template A constructs pair-conditioned directions; B and C evaluate them. The natural templates are those in Section~\ref{sec:parameter-routing}.

Let $g$ index a country pair. Let $x_{\tau,g,i}$ be the complete paired-country input for pair $g$, template $\tau\in\{A,B,C\}$, and requested position $i\in\{1,2\}$. The recorded vectors are $\vect{h}_{x_{\tau,g,1}}^{\ell}$ and $\vect{h}_{x_{\tau,g,2}}^{\ell}$. Template and requested position construct the input; they are not extra intervention conditions. For compact notation in this comparison, $\vect{\mu}^{\ell,\mathrm{first}}\equiv\vect{\mu}_{\mathrm{first}}^{\ell,\mathrm{main}}$ and $\vect{\mu}^{\ell,\mathrm{second}}\equiv\vect{\mu}_{\mathrm{second}}^{\ell,\mathrm{main}}$ are the respective averages of the three first- and three second-request states in the paired-task fitting set. Define a knowledge-and-marker exclusion function,
\begin{equation}
\mathtt{Exclude}(\vect{v};\vect{P}^{\ell},\vect{\eta}^{\ell})
=(\vect{I}_{H}-\vect{P}^{\ell})\vect{v}
-\vect{\eta}^{\ell}(\vect{\eta}^{\ell})^{\top}
(\vect{I}_{H}-\vect{P}^{\ell})\vect{v},
\label{eq:matched-exclusion}
\end{equation}
where $\vect{v}$ is an input state difference, $\vect{P}^{\ell}$ is the paired-task content projector, and $\vect{\eta}^{\ell}$ is the auxiliary marker contrast normalized after content exclusion. Both request constructions use the same two excluded objects:
\begin{equation}
\begin{aligned}
\vect{r}^{\ell,\mathrm{global}}
&=\mathtt{Unit}\!\left(\mathtt{Exclude}\!\left(
\vect{\mu}^{\ell,\mathrm{first}}-\vect{\mu}^{\ell,\mathrm{second}};
\vect{P}^{\ell},\vect{\eta}^{\ell}\right)\right),\\
\vect{r}_{g}^{\ell,\mathrm{pair}}
&=\mathtt{Unit}\!\left(\mathtt{Exclude}\!\left(
\vect{h}_{x_{A,g,1}}^{\ell}-\vect{h}_{x_{A,g,2}}^{\ell};
\vect{P}^{\ell},\vect{\eta}^{\ell}\right)\right).
\end{aligned}
\label{eq:matched-directions}
\end{equation}
The superscripts distinguish the shared fit direction from the country-pair direction. The two centers are
\begin{equation}
\vect{b}^{\ell,\mathrm{fit}}
=\tfrac12(\vect{\mu}^{\ell,\mathrm{first}}+\vect{\mu}^{\ell,\mathrm{second}}),
\qquad
\vect{b}_{g}^{\ell,\tau,\mathrm{pair}}
=\tfrac12(\vect{h}_{x_{\tau,g,1}}^{\ell}+\vect{h}_{x_{\tau,g,2}}^{\ell}).
\label{eq:matched-centers}
\end{equation}
Here $\vect{b}^{\ell,\mathrm{fit}}$ is the fit center $\vect{b}^{\ell}$ used in Section~\ref{sec:joint-process}, while $\vect{b}_{g}^{\ell,\tau,\mathrm{pair}}$ is the same pair-specific center denoted $\vect{b}_{x}^{\ell,\mathrm{pair}}$ when the evaluated input $x$ identifies that pair and template. Their four combinations define $c$ in Equation~\ref{eq:controlled-deletion}. Reversal doubles that equation's deletion vector. The natural comparison uses only its own pair-conditioned direction and pair center, with knowledge and auxiliary exclusion fitted in the natural-task coordinates. The auxiliary raw marker data are shared, but the content exclusion is task-specific.

\subsection{Aggregation, Uncertainty, and Actual Length}

Let $\mathcal{G}$ contain the same 24 validation country pairs denoted by $\mathcal{Q}$ in Appendix~\ref{app:expanded-evidence} and let $\mathcal{X}_g$ contain the four evaluated questions for pair $g$. For a fixed layer set $\mathcal{L}$, define the pair-averaged loss and the corresponding random-corrected effect:
\begin{equation}
\begin{aligned}
\overline{\Delta m}_{\mathcal{L},g}^{c}
&=\frac{1}{|\mathcal{L}|\,|\mathcal{X}_g|}
\sum_{\ell\in\mathcal{L}}\sum_{x\in\mathcal{X}_g}
\Delta m_{x}^{\ell,c},\\
E_{\mathcal{L},c}^{\mathrm{ans}}
&=\frac{1}{|\mathcal{G}|}\sum_{g\in\mathcal{G}}
\overline{\Delta m}_{\mathcal{L},g}^{c}
-\max_{j=1,\ldots,8}\frac{1}{|\mathcal{G}|}
\sum_{g\in\mathcal{G}}\overline{\Delta m}_{\mathcal{L},g}^{c,j}.
\end{aligned}
\label{eq:matched-aggregation}
\end{equation}
The overbar denotes averaging within a country pair and layer set; superscript $c,j$ denotes random control $j$ matched to intervention $c$. Thus the maximum is taken after, not before, averaging. For later knowledge, substitute the source-to-receiver knowledge-score loss from Equation~\ref{eq:single-query-later-knowledge}. All 24 pairs qualify in both tasks before intervention, so qualified-pair and all-pair estimates coincide.

The earlier layers are fixed at 29 and 31, the later set is $\mathcal{L}_{\mathrm{late}}=\{32,33,34,35,36\}$, and the source-to-receiver knowledge comparison is fixed at 32 to 36. Each bootstrap sample resamples 24 country pairs with replacement. The same sampled pairs are used across conditions, and the maximum random-control mean is recomputed within each sample before taking a condition difference. Intervals are the 2.5th and 97.5th percentiles of 2,000 samples; they are not simultaneous intervals across comparisons. Independent model reloads reproduce the same scientific fields and are not additional statistical samples.

\begin{table}[htbp]
\centering
\caption{Direct contrasts of the common-protocol late answer effects. Each row subtracts the second named condition from the first using shared country-pair bootstrap samples. Direction changes retain the stated center; center changes retain the stated direction.}
\label{tab:matched-contrasts}
\small
\begin{tabular}{lr}
\toprule
Comparison & Difference [95\% interval]\\
\midrule
Pair-conditioned minus global; fit center & $3.287$ [$2.674$, $3.821$]\\
Pair-conditioned minus global; pair center & $3.085$ [$2.481$, $3.629$]\\
Pair minus fit center; global direction & $0.161$ [$0.117$, $0.210$]\\
Pair minus fit center; pair-conditioned direction & $-0.041$ [$-0.082$, $0.005$]\\
Natural minus paired; pair direction and center & $-0.650$ [$-0.885$, $-0.395$]\\
\bottomrule
\end{tabular}
\end{table}

Matching each intervention to its own random controls does not equalize the true interventions. Across the 480 paired-task question--layer observations in the later set, mean actual deletion lengths are 16.38 for the global fit-centered condition and 80.47 for the pair-conditioned fit-centered condition. Hence Table~\ref{tab:matched-contrasts} identifies dependence on the deletion definition, including its projected length, not orientation alone. In addition, the global direction's earlier effect under this wording subset has an interval spanning zero. This does not isolate a precision effect relative to the older bfloat16 protocol, which also used different wording and order coverage.

Within the same paired-task content coordinates and fit center, layer-32 deletion produces corrected layer-36 knowledge losses of $-82.7$ [$-330.1$, $72.2$] for the global direction and $2648.4$ [$1223.3$, $3611.0$] for the pair-conditioned direction. Their direct difference is $2731.1$ [$1382.9$, $3691.5$]. These scores use Equation~\ref{eq:single-query-knowledge-score}; they are not percentages or cross-task measures of how much knowledge was retrieved.

\subsection{Random-Control Sensitivity of Natural-Question Knowledge Effects}
\label{app:random-control-sensitivity}

This is a post hoc sensitivity analysis of completed records, not a replacement of the registered natural-question test. For all 96 matched questions, baseline and edited answer margins and raw layer-32-to-36 deletion knowledge losses are identical between the two runs. The runs differ in their random controls. We pool their sixteen random controls, retaining the maximum-after-averaging rule and using a common country-pair bootstrap sequence for comparison.

\begin{table}[htbp]
\centering
\caption{Sensitivity of the natural-question layer-32-to-36 deletion knowledge effect to the random controls. Raw true deletion losses are identical. The common bootstrap sequence can produce slightly different interval endpoints from the original registered report.}
\label{tab:random-control-sensitivity}
\small
\begin{tabular}{lr}
\toprule
Random controls & Knowledge effect [95\% interval]\\
\midrule
Original eight & $1102.5$ [$-47.6$, $2128.2$]\\
Additional eight & $1376.1$ [$184.6$, $2292.5$]\\
Pooled sixteen & $1102.5$ [$-50.3$, $2127.8$]\\
\bottomrule
\end{tabular}
\end{table}

Pooling the controls retains an inconclusive deletion knowledge effect (Table~\ref{tab:random-control-sensitivity}). The pooled reversal effect remains positive at $6270.8$ [$3261.0$, $8747.7$]. A favorable set of eight controls therefore does not justify upgrading deletion necessity, although the directional reversal finding remains supported.

\subsection{Direct-Selection Readouts and Intervention}
\label{app:direct-selection}

This experiment measures direct object selection in the same Qwen instruction weights and continent-name task, using float32. The original six fitting countries form all 24 ordered cross-continent pairs, each with two requests under template A, giving $N_{\mathrm{fit}}=48$ fitting questions. Reordering these countries adds configurations, not independent country identities. The original three selection pairs are evaluated in both orders and under templates B and C. All requests are checked for baseline correctness, but source screening changes first to second only, giving $N_{\mathrm{sel}}=12$ screening questions.

Four target outputs are fitted separately: the identities of both countries, the first country's continent, the second country's continent, and the requested first/second label. Country-identity targets concatenate the first five principal-component coordinates of each country's mean input-token embedding, fitted only on the six fitting countries. Each candidate continent uses a three-dimensional one-hot target independent of which country is requested. The selection target is the scalar $+1$ or $-1$. The relation is always continent, so identity preservation does not measure preservation of arbitrary relation intent.

For each layer, subtract the mean fit state $\vect{\mu}^{\ell}$ and divide by the scalar $\zeta^{\ell}$, the root-mean-square value of all centered fitting-state coordinates. Write the standardized state as $\widetilde{\vect{h}}_{x}^{\ell}=(\vect{h}_{x}^{\ell}-\vect{\mu}^{\ell})/\zeta^{\ell}$. For each output component $c$, its target vector is $\vect{y}_x^c$, with the selection scalar treated as a one-element vector during fitting. We fit a weight matrix $\vect{W}^{\ell,c}$ and bias vector $\vect{\beta}^{\ell,c}$ by ridge regression, which adds a squared-weight penalty to least squares \citep{hastie2020ridge}:
\begin{equation}
(\vect{W}^{\ell,c},\vect{\beta}^{\ell,c})
=\mathop{\arg\min}_{\vect{W},\vect{\beta}}
\left\{
\frac{1}{N_{\mathrm{fit}}}\sum_{x\in\mathcal{X}_{\mathrm{fit}}}
\|\vect{W}^{\top}\widetilde{\vect{h}}_{x}^{\ell}+\vect{\beta}-\vect{y}_x^c\|_2^2
+0.1\|\vect{W}\|_{\mathrm{F}}^2
\right\},
\label{eq:direct-readout-fit}
\end{equation}
where $\mathcal{X}_{\mathrm{fit}}$ is the set of fitting questions and the squared Frobenius norm sums the squared weight entries \citep[Section~3.1]{halko2011structure}. The bias is unpenalized. No validation result selects the regularization coefficient.

Define the readout function
\begin{equation}
\mathtt{Read}(\vect{h};\vect{W},\vect{\beta},\vect{\mu},\zeta)
=\vect{W}^{\top}(\vect{h}-\vect{\mu})/\zeta+\vect{\beta}.
\label{eq:direct-readout}
\end{equation}
Its output at layer $\ell$ for question $x$ under intervention $e$ is $\widehat{\vect{y}}_{x}^{\ell,c,e}$, using the fitted parameters for component $c$ at that layer. For selection, $\widehat{y}_{x}^{\ell,\mathrm{sel},e}$ denotes the sole scalar coordinate. Unit-normalizing the selection weight gives $\vect{o}^{\ell,\mathrm{direct}}$ in Equation~\ref{eq:direct-selection-swap}; rescaling the weight to original state units does not change that unit direction.

At each source layer, two random controls are generated from Gaussian linear combinations of centered fitting states, projected orthogonally to $\vect{o}^{\ell,\mathrm{direct}}$, and scaled to the true swap's actual written length. A wrong-position control applies the same swap vector at the first token of ``continent'' in the question. No answer association is supplied in these inputs.

\subsection{Joint Criterion and Prespecified Stopping}

For an identity or candidate-content component $c$, let
\begin{equation}
\nu_{x}^{\ell',c,e}
=\|\widehat{\vect{y}}_{x}^{\ell',c,e}
-\widehat{\vect{y}}_{x}^{\ell',c,0}\|_2
\label{eq:direct-content-change}
\end{equation}
be its readout change relative to the unmodified state. The indicator $\chi_{x}^{\ell'}$ is one only if the swap's change in each of the three content outputs is at most the larger of that output's two random-control changes, plus a numerical tolerance of $10^{-8}$. It is zero otherwise. The source-layer indicator is $\chi_{x}^{\ell}$; receiving layers satisfy $\ell'>\ell$.

Let $\sigma_x=+1$ for a first-object request and $-1$ for a second-object request. The directed selection-prediction change is
\begin{equation}
\omega_{x}^{\ell',e}
=-\frac{\sigma_x}{2}
\left(\widehat{y}_{x}^{\ell',\mathrm{sel},e}
-\widehat{y}_{x}^{\ell',\mathrm{sel},0}\right).
\label{eq:direct-selection-change}
\end{equation}
The denominator is the distance between the target labels $+1$ and $-1$, not a knowledge quantity. Positive changes move the prediction toward the opposite request. Let $\omega_{x}^{\ell',\mathrm{ctrl}}$ and $\Delta m_x^{\mathrm{ctrl}}$ be, respectively, the largest directed selection change and largest answer-margin loss among the two random controls and the wrong-position control. These maxima are taken separately for each question and outcome.

The screening answer effect and receiving-layer joint fraction are
\begin{equation}
E_{\mathrm{swap}}^{\ell,\mathrm{ans,pointwise}}
=\frac{1}{N_{\mathrm{sel}}}\sum_{x\in\mathcal{X}_{\mathrm{sel}}}
(\Delta m_x^{\mathrm{swap}}-\Delta m_x^{\mathrm{ctrl}}),
\label{eq:direct-answer-effect}
\end{equation}
\begin{equation}
\begin{aligned}
F^{\ell\rightarrow\ell',\mathrm{joint}}
=\frac{1}{N_{\mathrm{sel}}}\sum_{x\in\mathcal{X}_{\mathrm{sel}}}
&\chi_{x}^{\ell}\chi_{x}^{\ell'}\,
\mathtt{Ind}(\omega_{x}^{\ell',\mathrm{swap}}-\omega_{x}^{\ell',\mathrm{ctrl}}\geq0.25)\\[-1mm]
&\cdot\mathtt{Ind}(\Delta m_x^{\mathrm{swap}}-\Delta m_x^{\mathrm{ctrl}}>0),
\end{aligned}
\label{eq:direct-joint-fraction}
\end{equation}
where $\mathcal{X}_{\mathrm{sel}}$ contains the 12 screening questions and $\mathtt{Ind}$ is the previously defined indicator. The superscript pointwise marks the maximum-before-averaging rule in Equation~\ref{eq:direct-answer-effect}; it differs from Equation~\ref{eq:matched-aggregation}.

A source qualifies only if mean source content preservation is at least 0.70, mean source selection change is at least 0.25, Equation~\ref{eq:direct-answer-effect} is positive, and at least three receiving layers have $F^{\ell\rightarrow\ell',\mathrm{joint}}\geq0.60$. Numerical qualification additionally requires at least 95\% of active swap cases to meet the 5\% target-write and random-length tolerances. Sources 1--33 are screened so that at least three later layers exist. The best source and three receiving layers would then be frozen before validation, using mean joint fraction, answer effect, and earlier layer as successive tie-breakers. No source meets these conditions.

The 24 validation pairs were presplit into groups of twelve. Validation and restoration would use eight covariance-based random controls and a wrong-position control. Both groups would have to pass the fixed source-preservation and three-receiver criteria before two continuations from the same selection-modified state could test both candidate answers. Baseline correctness and any restoration success require the intended answer to score above both other members of $\mathcal{K}$ under Equation~\ref{eq:single-query-answer-margin}; directional effects still compare the two objects' answers. All pairs pass baseline qualification, but validation interventions and restoration are not run because screening fails.

Each model load produces 372 baseline records and 396 screening records, with identical scientific fields between loads. Every load's 1,584 actual writes passes the 5\% numerical tolerance, including checks for a nonzero target collapsing to zero. This isolates the reported boundary from those checked numerical failures, but the finite linear readouts remain an incomplete measure of identity and candidate knowledge. No inference that selection is absent from the model follows.

\subsection{Base-Model Capital Selection and Two-Continuation Restoration}
\label{app:base-capital-selection}

The separate positive result in Section~\ref{sec:object-selection} uses Qwen2.5-3B base weights and the capital relation. Each prompt names two countries but supplies neither capital. Two held-out phrasings have the forms
\begin{quote}
\small
Consider Argentina and Canada. What is the capital city of the first one?\\
Answer with only the name.\\[1mm]
Argentina; Canada. Give the capital city of the second listed place.\\
Answer with only the name.
\end{quote}
Changing \emph{first} to \emph{second} changes the requested object while leaving the two named countries and the relation fixed. The model must produce one of the two capitals from its weights.

Forty-four country pairs pass the behavioral gate. Fifteen pairs fit four separate linear interfaces. The shared two-country query target concatenates the fit-only principal-component coordinates of the first and second country-name input embeddings. The two candidate-capital targets are the corresponding fit-only coordinates of the first and second capital-name embeddings. These three targets are unchanged when the requested object changes. The fourth target is a two-entry selection label, $(1,0)^\top$ for the first object and $(0,1)^\top$ for the second. The intervention and restoration use the selection subspace fitted to these two outputs; this is distinct from the scalar $+1/-1$ readout in the instruction-model tests. Eight disjoint pairs screen all 36 layers and five registered input positions. Before either held-out group is opened, this scan fixes decoder layer 26 at the question-end position as the source. Validation A contains ten pairs and 40 questions; validation B contains eleven pairs and 44 questions. Each group combines two phrasings with both requested objects.

At the fixed source, the intervention replaces only the fitted selected-object component with its value under the opposite request. Eight matched random changes and an application of the same change at the wrong input position serve as controls. A validation question passes the first stage only if the selected-object readout moves toward the other request, the answer margin moves toward the other capital beyond every control, and the shared-query plus both candidate-capital readouts remain within the matched-random bounds. The downstream selected-object change exceeds all controls at layers 27--30. The answer criterion passes on 37/40 questions in validation A and 44/44 in validation B; the other three readouts are preserved on every validation question.

The final test starts twice from the \emph{same} source-modified state. At a receiving layer, let the changed state be $\vect{h}^{\mathrm{chg}}$, and let $\vect{h}^{\mathrm{src}}$ and $\vect{h}^{\mathrm{opp}}$ be the clean states for the original and opposite requests. Each branch adds only the projection of $\vect{h}^{\mathrm{src}}-\vect{h}^{\mathrm{chg}}$ or $\vect{h}^{\mathrm{opp}}-\vect{h}^{\mathrm{chg}}$ into that receiver's fitted selection basis. Thus one continuation restores the clean original-request selection coordinate and the other restores the clean opposite-request coordinate without copying the full clean state. A question counts only if the two continuations respectively favor the two capital answers beyond eight random restorations and the wrong-position restoration, while all three non-selection readouts remain within random bounds. The joint pass counts at layers 27, 28, and 29 are 25/40, 28/40, and 24/40 in validation A, and 34/44, 34/44, and 32/44 in validation B. Both groups therefore meet the prespecified 60\% threshold at the same three receivers. Layer 30 is excluded because validation A reaches only 23/40, or 57.5\%, on the same two-answer criterion.

An earlier control edits the hidden state at the ordinal word's token position in the first decoder block, recorded as zero-based layer 0. It changes later selection and answers, but restoring the original request at layers 15--19 recovers the original answer on at most 2/40 and 1/44 questions. The early positional edit therefore does not meet the same two-usable-candidate definition. The confirmed result is limited to the layer-26 question-end source and layers 27--29 in this Qwen base-model capital task.

\subsection{Capital Replication in Three Instruction Models}
\label{app:cross-model-capital-selection}

We next keep the capital relation, phrasings, country-disjoint splits, scoring rule, and functional gates fixed while replacing the base weights with Qwen-2.5-3B-Instruct, Llama-3.2-3B-Instruct, or Gemma-3-4B-Instruct. Each model uses its own chat template and decoder depth, and all hidden-state calculations use float32. The task contains 196 questions. Qwen answers 195 correctly; Llama and Gemma answer all 196. The resulting qualified counts are 15 fit pairs, seven Qwen or eight Llama/Gemma screening pairs, and 13 pairs in each held-out group.

At each layer, ridge regression \citep{hastie2020ridge} with coefficient 0.1 fits four readouts using only the fit questions, following Equation~\ref{eq:direct-readout-fit}. The targets are the concatenated identities of both countries, the first capital, the second capital, and the scalar first/second selection label. Country and capital targets are coordinates of their mean input-token embeddings in fit-only principal-component spaces; no held-out result chooses the rank or coefficient. Across all three models, the 15 country targets and 15 capital targets each have rank 14 with no coincident target pair.

Screening applies Equation~\ref{eq:direct-selection-swap} at the question-end position of every possible source layer. Two covariance-based random directions orthogonal to the fitted selection direction and a wrong-position application control each question. A source must preserve the other three readouts on at least 70\% of screening questions, move the source selection score by at least 0.25, have positive pointwise-corrected answer effect, and produce at least three receivers with $F^{\ell\rightarrow\ell',\mathrm{joint}}\geq0.60$. The selected source and three receivers are then frozen. Each held-out group contains 13 pairs, two phrasings, and both requested objects, giving 52 questions; validation uses eight covariance controls plus the wrong-position control. Two-candidate restoration remains locked unless both groups pass all three fixed receivers.

\begin{table}[htbp]
\centering
\caption{Direct capital-selection outcomes in the three instruction models. Screening counts use 14 questions for Qwen and 16 for Llama and Gemma. Each held-out group contains 52 questions. A dash means that the prespecified earlier gate failed, so the later stage was not run.}
\label{tab:cross-model-capital-selection}
\small
\begin{tabular}{llll}
\toprule
Model & Screening outcome & Held-out outcome & Restoration\\
\midrule
Qwen & No qualifying source & --- & Not run\\
Llama & Source 14; only two qualifying receivers & --- & Not run\\
Gemma & Source 17; receivers 18--20 fixed & Three-receiver gate fails & Not run\\
\bottomrule
\end{tabular}
\end{table}

Qwen's most informative content-preserving candidate, source layer 21, preserves all three non-selection readouts on 10/14 questions and moves the fitted selection coordinate by 0.995 of the full first-to-second target difference. Its mean answer change is nevertheless 0.008 answer-margin units \emph{below} the strongest pointwise control, and its best receiving-layer joint count is 5/14. No source qualifies.

Llama source layer 14 preserves the other readouts on all 16 screening questions. Its selection change is 1.007 of the full target difference, and its mean answer change exceeds the strongest pointwise controls by 1.010 answer-margin units. The joint counts are 14/16 at layer 15, 11/16 at layer 16, and 9/16 at layer 17. Because 60\% of 16 questions requires ten passes, only two receivers qualify and held-out validation is not opened.

Gemma source layer 17 passes screening: source content is preserved on 14/16 questions, its selection change is 0.977 of the full target difference, and its pointwise-corrected answer effect is 1.841. The frozen receivers are layers 18, 19, and 20. In both 52-question held-out groups, source content is preserved on all questions and the selection change at every fixed receiver exceeds all controls on all questions. The answer change exceeds its strongest pointwise control on 32/52 questions in validation A and 31/52 in validation B. The resulting joint counts are 32, 32, and 31 in validation A and 31, 31, and 29 in validation B. Since every fixed receiver must reach 60\%, equivalent to 32/52 questions, validation A fails at receiver layer 20 and validation B fails at all three receivers. Both groups therefore fail the gate, and restoration remains locked.

Primary and fresh-load runs agree record by record for every model, and every active write passes the registered numerical tolerances. The Gemma result therefore identifies a reproducible held-out internal selection candidate whose answer transmission is near the threshold; it does not establish that two capital facts remain separately usable from one modified state. No instruction model adds a strict object-selection positive result.

\FloatBarrier
\section{Natural-Question Wording and Complete Country Lists}
\label{app:natural-inputs}

This appendix specifies every question used in Section~\ref{sec:parameter-routing}. The task is always to name the continent of one country. The tables provide the complete country pools, and the three templates below determine the full question text for each country. Pairing is an analysis construction: each country is submitted in a separate model input.

\subsection{Question Wording and Input Construction}

For a country $s$, the raw question text $p_{\tau,s}$ uses one of the following wordings, followed on a new line by the instruction \emph{Answer with only the continent name.}
\begin{center}
\small
\begin{tabular}{lp{0.82\linewidth}}
\toprule
Template & Question text\\
\midrule
A & Which continent is \{country\} located in?\\
B & On which continent is \{country\} located?\\
C & What continent is \{country\} in?\\
\bottomrule
\end{tabular}
\end{center}
The country name replaces \{country\}. The checkpoint's chat template then adds user and assistant-generation markers, and its tokenizer converts the formatted text to $x_{\tau,s}$. Thus the final input position is determined after formatting and tokenization, not by locating the final printed word \emph{name}. The model and tokenizer identities and rendered-input hashes are retained in the frozen experiment records.

Template A supplies one state per fitting country for the content references and supplies the paired states used to construct each request direction. Templates B and C supply the intervention-evaluation questions. There are six content-fitting inputs, 54 direction-construction inputs from the 27 screening and validation pairs, and 108 evaluated questions (12 screening and 96 validation). The complete 168 raw question texts and their roles are included in \path{anc/evidence/natural_question_inputs/questions.jsonl}; the file contains questions, not answer strings appended as factual context.

\subsection{Fitting Countries}

The content fitting set contains exactly two countries per continent:
\begin{center}
\small
\begin{tabular}{lll}
\toprule
Continent & Country 1 & Country 2\\
\midrule
Africa & Kenya & Nigeria\\
Asia & China & Thailand\\
Europe & France & Spain\\
\bottomrule
\end{tabular}
\end{center}
For each country, only the template-A question contributes a vector to the fitted mean. Consequently, the full mean averages six vectors and each continent mean averages two. No screening or validation country contributes to these content fits.

\subsection{Screening and Validation Pairs}

The following three pairs select the source and receiving layers. Each pair contributes two countries under B and C, giving four evaluated questions:
\begin{center}
\small
\begin{tabular}{llll}
\toprule
First country & Continent & Second country & Continent\\
\midrule
Egypt & Africa & India & Asia \\
Vietnam & Asia & Germany & Europe \\
Portugal & Europe & Morocco & Africa \\
\bottomrule
\end{tabular}
\end{center}

The 24 disjoint validation pairs below test the fixed choice. No country in this table occurs in the fitting or screening lists:
\begin{center}
\small
\begin{tabular}{llll}
\toprule
First country & Continent & Second country & Continent\\
\midrule
Angola & Africa & Japan & Asia \\
Cambodia & Asia & Hungary & Europe \\
Albania & Europe & Gabon & Africa \\
Benin & Africa & South Korea & Asia \\
Malaysia & Asia & Iceland & Europe \\
Austria & Europe & Malawi & Africa \\
Botswana & Africa & North Korea & Asia \\
Philippines & Asia & Ireland & Europe \\
Belgium & Europe & Mali & Africa \\
Burkina Faso & Africa & Nepal & Asia \\
Brunei & Asia & Italy & Europe \\
Bulgaria & Europe & Mozambique & Africa \\
Burundi & Africa & Bhutan & Asia \\
Maldives & Asia & Netherlands & Europe \\
Croatia & Europe & Namibia & Africa \\
Cameroon & Africa & Pakistan & Asia \\
Qatar & Asia & Norway & Europe \\
Czechia & Europe & Niger & Africa \\
Chad & Africa & Sri Lanka & Asia \\
Kuwait & Asia & Poland & Europe \\
Finland & Europe & Rwanda & Africa \\
Ethiopia & Africa & Laos & Asia \\
Oman & Asia & Romania & Europe \\
Greece & Europe & Senegal & Africa \\
\bottomrule
\end{tabular}
\end{center}
Each validation pair again supplies four evaluated questions. The pair, with all its wordings and intervention conditions kept together, is the independent resampling unit. A and B/C refer to different construction and evaluation roles, not additional model inputs supplied alongside a question.

\FloatBarrier
\end{document}